\documentclass[letterpaper]{article} %
\usepackage[preprint]{aaai2027}%
\usepackage[hyphens]{url}  %
\usepackage{graphicx} %
\usepackage{natbib}  %
\usepackage{caption} %
\usepackage{algorithm}
\usepackage{algorithmic}

\usepackage{newfloat}
\usepackage{listings}
\usepackage{amsthm}
\usepackage{multirow} 
\usepackage[most]{tcolorbox}
\usepackage{subcaption}
\DeclareCaptionStyle{ruled}{labelfont=normalfont,labelsep=colon,strut=off} %
\floatstyle{ruled}
\newfloat{listing}{tb}{lst}{}
\floatname{listing}{Listing}

\usepackage{booktabs}

\usepackage{amsmath}
\usepackage{bm}

\usepackage{amssymb}
\usepackage[table]{xcolor} %
\newcommand{\joseph}[1]{\textcolor{black}{#1}}

\newcommand{\gianni}[1]{\textcolor{black}{#1}}

\newcommand{\method}{SGPU} %

\def\vtheta{{\bm{\theta}}}

\def\vdelta{{\bm{\delta}}}

\def\vphi{{\bm{\phi}}}
\def\vlambda{{\bm{\lambda}}}
\def\mDelta{{\bm{\Delta}}}

\def\eqref#1{equation~\ref{#1}}

\def\1{\bm{1}}

\def\vmu{{\bm{\mu}}}
\def\vtheta{{\bm{\theta}}}

\def\ve{{\bm{e}}}
\def\vf{{\bm{f}}}

\def\vl{{\bm{l}}}

\def\vu{{\bm{u}}}
\def\vv{{\bm{v}}}

\def\vx{{\bm{x}}}
\def\vy{{\bm{y}}}

\def\mD{{\bm{D}}}

\def\mI{{\bm{I}}}

\def\mK{{\bm{K}}}

\def\mP{{\bm{P}}}

\def\mU{{\bm{U}}}

\def\mX{{\bm{X}}}

\def\mPhi{{\bm{\Phi}}}
\def\mLambda{{\bm{\Lambda}}}
\def\mSigma{{\bm{\Sigma}}}

\DeclareMathAlphabet{\mathsfit}{\encodingdefault}{\sfdefault}{m}{sl}
\SetMathAlphabet{\mathsfit}{bold}{\encodingdefault}{\sfdefault}{bx}{n}

\newcommand{\E}{\mathbb{E}}

\newcommand{\R}{\mathbb{R}}

\DeclareMathOperator{\Tr}{Tr}

\newtheorem{lemma}{Lemma}
\newtheorem{assumption}{Assumption}
\newtheorem{remark}{Remark}
\usepackage{comment}
\title{Improving Semantic Uncertainty Quantification in LVLMs  \\ with Semantic Gaussian Processes}
\author{
    Joseph Hoche\textsuperscript{\rm 1},
    Andrei Bursuc\textsuperscript{\rm 2},
    David Brellmann\textsuperscript{\rm 3},
    Gilles Louppe\textsuperscript{\rm 4},\\
    Pavel Izmailov\textsuperscript{\rm 5},
    Angela Yao\textsuperscript{\rm 6},
    Gianni Franchi\textsuperscript{\rm 1,7}
}
\affiliations{
    \textsuperscript{\rm 1}AMIAD, Pôle Recherche, Palaiseau\\
    \textsuperscript{\rm 2}valeo.ai\quad
    \textsuperscript{\rm 3}Safran Tech\quad
    \textsuperscript{\rm 4}University of Liège\\
    \textsuperscript{\rm 5}New York University\quad
    \textsuperscript{\rm 6}National University of Singapore\quad
    \textsuperscript{\rm 7}ENSTA Paris\\
    
}

\begin{document}

\maketitle

\begin{abstract}

Large Vision-Language Models (LVLMs) often produce plausible but unreliable outputs, making uncertainty estimation essential. 
Recent work on uncertainty estimates compare multiple sampled responses in a semantic space to measure their consistency. 
However, such comparisons are often fragile. 
They are sensitive to minor phrasing variations, and may fail to distinguish semantically distinct answers, making the uncertainty estimates unreliable.
To address these limitations, we propose \textbf{Semantic Gaussian Process Uncertainty (SGPU)}.  SGPU is a Bayesian framework that maps semantic consistency patterns in generated answers to a predictive uncertainty.
SGPU maps generated answers into a dense semantic space, computes the Gram matrix of their embeddings, and summarizes their semantic configuration via the eigenspectrum. This spectral representation is then fed into a Gaussian Process Classifier \gianni{that predicts whether the sampled answers are likely to be correct} and that can be applied in both black-box and white-box settings.
Across six LLMs and LVLMs on eight datasets spanning VQA, image classification, and textual QA, \gianni{SGPU achieves the best aggregate performance compared with both training-free and supervised uncertainty-estimation methods.}
We further show that SGPU transfers across models and modalities, requiring minimal tuning and limited training samples. 
Code and data will be fully available upon acceptance.

\end{abstract}

\section{Introduction}
\label{sec:intro}

Large Vision-Language Models (LVLMs) extend Large Language Models (LLMs) to multimodal inputs and excel at tasks requiring visual and language understanding~\cite{openai2024gpt4, geminiteam2024}.
Despite their success, LVLMs are prone to hallucinations: they can produce plausible but incorrect outputs that are unfaithful to the input~\cite{ji2023, liu2024survey}.
This raises the need for uncertainty quantification (UQ) to detect when generated content should not be trusted~\cite{hernandez2015, gal2016dropout}.

Initially, UQ methods for LVLMs were \gianni{mainly based on token probabilities}.
They operate directly on autoregressive token probabilities, using measures such as log-perplexity or predictive entropy~\cite{malinin2021,aichberger2024rethinking}.
For instance, given an image of the Eiffel Tower and the question \emph{``In which city can I find this monument?''}, an LVLM may generate several answers that all refer to Paris, such as \emph{``The figure shows the Eiffel Tower in Paris.''} and \emph{``The Eiffel Tower is depicted in this image.''}.
Token-based scores may assign different uncertainty values to these phrases because they use different words and sentence structures, even though they have the same meaning~\cite{aichberger2024rethinking}.

Semantic uncertainty methods \gianni{solve it by comparing several sampled responses in a semantic space}~\cite{farquhar2024detecting}.
This is typically done using embeddings derived from the hidden states of LVLMs~\cite{chen2024,binkowski2025},
or from external models such as Natural Language Inference (NLI) systems~\cite{farquhar2024detecting, nikitin2024kernel},
instruction-tuned LLMs~\cite{farquhar2024detecting, nikitin2024kernel, ji2025},
or sentence encoders~\cite{grewal2024improving, abdaljalil2025}.
When the model produces conflicting answers, such as \textit{Paris}, \textit{Las Vegas}, and \textit{Tokyo}, semantic uncertainty should be high.

In practice, however, semantic comparisons can be fragile.
For instance, NLI models are highly sensitive to small changes in phrasing, additional correct details, or uninformative words that can change their similarity predictions~\cite{arakelyan2024, grewal2024improving}.
Similarly, representations extracted from the hidden states of LVLMs are not explicitly trained to capture all the semantic properties of complete generated answers.
\gianni{
They also require white-box access to the evaluated model, which prevents their use with closed-source models or external APIs. 
}
Equivalent answers may receive low similarity scores, while answers with different meanings may appear similar.
As a result, semantic uncertainty estimates can depend strongly on the representation, similarity function, clustering method, and associated hyperparameters.

\gianni{Another important distinction concerns supervision.
Most semantic uncertainty methods directly transform an entropy, similarity, or spectral statistic into an uncertainty score and do not require labelled data
but a calibration set to tune the parameters.
In contrast, supervised methods learn to predict correctness from labelled examples.
Although supervised approaches can improve performance, existing methods such as SAPLMA~\citep{azaria2023the} and Semantic Entropy Probes~\citep{kossen2024semantic} rely on model-specific internal representations.
They require retraining or adaptation when the underlying LLM or LVLM changes and otherwise generalize poorly within the same model~\citep{orgad2025llms}}.

\gianni{
These limitations motivate the following research question:
\emph{How can we reliably estimate whether responses generated by an LLM or LVLM are trustworthy, using only multiple sampled answers, while remaining applicable to black-box models and transferable across architectures?}
A useful solution should capture the meaning of complete answers, require only limited labelled data, and avoid model-specific internal representations so that it can be reused across different models and modalities.
}

\begin{figure*}[t]

    \centering
    \includegraphics[width=1\textwidth]{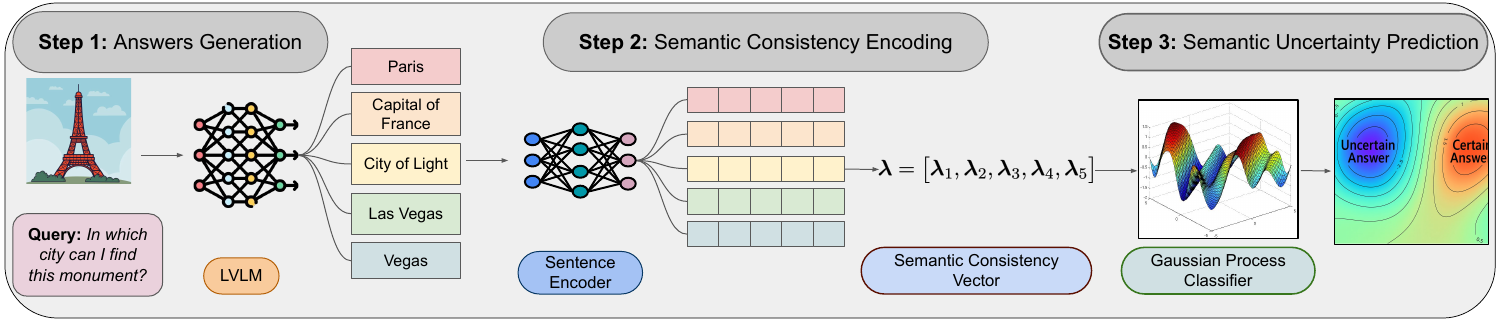}
    \caption{\textbf{Illustration of the SGPU pipeline.} SGPU operates in three steps: (1) generating multiple candidate sequences for a given context (query + image), (2) encoding their semantic consistency into a single vector representation $\bm{\lambda}$  (Section~\ref{sec:SGPU-embeddings}), and (3) using this vector representation as input for a Gaussian Process Classifier to estimate the predictive semantic uncertainty / truthfulness of the generated samples (Section~\ref{sec:SGPU-GP}).}
    \label{fig:pipeline0}

\end{figure*}

To address this question, we introduce \emph{Semantic Gaussian Process Uncertainty} (SGPU).
SGPU is a supervised Bayesian framework that learns to map semantic consistency patterns among generated answers to the probability that the response set is trustworthy; see Figure~\ref{fig:pipeline0}.
Given an input containing text, an image, or both, we sample multiple answers from an LLM or LVLM and embed each answer using an external sentence encoder trained to capture sentence-level semantics.

We then construct a Gram matrix from these embeddings and summarize the semantic configuration of the answers using its eigenspectrum.
\gianni{
As shown in our work, the eigenspectrum provides a compact description of how the generated answers are distributed across different meanings.
}
SGPU feeds this spectral representation into a \emph{Gaussian Process Classifier} (GPC), which learns to map semantic consistency patterns to predictive confidence.
\gianni{
Because its representation is computed only from generated answers using an external encoder, SGPU does not require access to hidden states, model parameters, or token probabilities.
It can therefore be applied to black-box models and transferred across architectures and tasks, unlike supervised techniques based on model-specific internal representations.
The GPC also provides an estimate of its own uncertainty and can be trained with a limited number of labelled examples.
}

\paragraph{Contributions.}
This work makes the following key contributions:
\gianni{
\textbf{(1) Spectral representation.}
We show that the eigenvalues of the answer Gram matrix describe how generated answers are distributed across meanings.
We also analyze how embedding errors affect this representation.
\textbf{(2) Bayesian trustworthiness prediction.}
We introduce SGPU, which uses the eigenspectrum of generated-answer embeddings as input to a Gaussian Process Classifier.
SGPU works without access to model parameters, hidden states, or token probabilities.
} To the best of our knowledge, this is the first approach to estimate LLM/LVLM confidence from multiple generated answers using a learned predictor (GPC). 
\gianni{
\textbf{(3) Comparison with supervised and training-free methods.}
SGPU achieves the best average AUROC, AUARC, and ECE across the evaluated settings and remains effective with only 200 labelled examples.
}
\gianni{
\textbf{(4) Transferability.}
Unlike supervised methods based on model-specific hidden states, SGPU can be applied without fine-tuning to unseen models and modalities.
}

\section{Related Works}
\label{sec:Related}

\paragraph{Bayesian Uncertainty Estimation.}

\gianni{Uncertainty quantification is a central topic in Bayesian machine learning~\cite{murphy2012machine}. Classical approaches include Gaussian Processes~\cite{williams2006gaussian} and Bayesian Neural Networks~\cite{mackay1992bayesian,mackay1995probable}, which place distributions over model parameters to represent predictive uncertainty. Scalable approximations have made Bayesian inference more practical for deep networks~\cite{blundell2015weight,gal2016dropout,laurent2023symmetry}, although their computational cost remains high for large vision models, motivating more efficient methods~\cite{maddox2019simple,franchi2020tradi}. Related alternatives include deep ensembles~\cite{laurent2022packed,gabetni2025ensembling} and Laplace approximations~\cite{ritter2018scalable,daxberger2021laplace}, which can provide accurate and well-calibrated uncertainty estimates~\cite{baumann2024post}. However, most of these approaches target non-generative architectures and remain difficult to apply to large LLMs and LVLMs because of their scale, multimodal structure, and complex output spaces.
}

\paragraph{Uncertainty Quantification in LLMs and LVLMs.}
\gianni{Recent UQ methods for LLMs and LVLMs include self-verbalized confidence~\cite{kadavath2022,lin2022teaching}, single-generation scores based on token probabilities or internal representations~\cite{malinin2021,murray2018,ren2023,azaria2023the,li2023}, perturbation-based approaches~\cite{gao2024,zhang2025,li2025semantic}, and semantic-consistency methods that compare multiple generated responses using external models, internal features, or confidence scores~\cite{malinin2021,kuhnsemantic,farquhar2024detecting,lin2024generating}. Our method belongs to the last category and estimates semantic uncertainty from response consistency using an external sentence encoder.
}

\section{Preliminaries}
\label{sec:background}
Given an input $\vx$ which may contain images and/or text, an LVLM parameterized by $\vtheta$ autoregressively generates an output response sequence of tokens $\vy=\bigl\{\vy_t\bigr\}_{t=1}^T$, where $\vy_t$ denotes the $t^{\text{-th}}$ output token and $T$ is the length of the output sequence.
The probability of the generated sequence $\vy$ is the joint probability of tokens, defined as the product of conditional token probabilities:
\begin{align}
    \begin{split}
    p(\vy \mid \vx, \vtheta)=\prod_{t=1}^T p(\vy_t \mid \vx, \vy_{<t}, \vtheta), \quad  \\ \text{where } \vy_{<t}=\bigl\{\vy_1, \ldots, \vy_{t-1}\bigr\}.
    \end{split}
\end{align}

\subsection{Uncertainty Estimation}
\label{sec:uq_estimation}
\paragraph{Log-Perplexity.} 
A straightforward uncertainty measure is the average negative log-likelihood of generated tokens or the log-perplexity~\cite{malinin2021, murray2018}. 
Log-perplexity is defined as the arithmetic mean of the log-probabilities of the generated tokens:
\(
        \log \operatorname{PPL}(\vy \mid \vx, \vtheta) = -\tfrac{1}{T} \sum_{t=1}^T \log p(\vy_t \mid \vx, \vy_{<t}, \vtheta)   =-\tfrac{1}{T}\log p(\vy \mid \vx, \vtheta).
\)

\paragraph{Predictive Entropy.} 
Another %
baseline %
is predictive entropy~\cite{farquhar2024detecting, lin2024generating, kadavath2022}.
Given an input $\vx$, predictive entropy is defined as the mean log-perplexity computed over a set of $N$ generated candidate sequences $\mathcal{Y}:=\{\vy^{(1)}, \ldots, \vy^{(N)}\}$:
\(
    H(\mathcal{Y} \mid \vx, \vtheta) = \tfrac{1}{N} \sum_{i=1}^N \log \operatorname{PPL}(\vy^{(i)} \mid \vx, \vtheta).
\)
$H(\mathcal{Y} \mid \vx, \vtheta)$ provides a Monte-Carlo approximation of the Shannon entropy $\E_{\vy}\bigl[-\log p(\vy \mid \vx, \vtheta)\bigr]$ over outputs generated by the LVLM.
The log-perplexity and the predictive entropy reach their highest values when the output is uncertain within the token space.

\paragraph{Semantic Entropy.} 
Given an input $\vx$, semantic entropy is defined over a set of generated sequences $\mathcal{Y}:=\{\vy^{(1)}, \ldots, \vy^{(N)}\}$ that are partitioned into semantically consistent clusters $\mathcal{C}:=\{\mathcal{C}_k\}_{k=1}^K$, such that sequences within each cluster share the same meaning.
In practice, clustering is performed either by using bidirectional entailment predictions from NLI models~\cite{kuhnsemantic, farquhar2024detecting, nikitin2024kernel} and instruction-tuned LLMs~\cite{farquhar2024detecting, nikitin2024kernel, ji2025} or by using embeddings derived from hidden states of the LVLM~\cite{lee2025} and external models~\cite{abdaljalil2025}.
The probability mass $p(\mathcal{C}_k \mid \vx, \vtheta)$ assigned to each cluster $\mathcal{C}_k$ is defined as the sum of normalized sequence probabilities of all outputs assigned to that cluster:
\begin{equation}
    p(\mathcal{C}_k \mid \vx, \vtheta) = \sum_{\vy \in \mathcal{C}_k} \tilde p(\vy \mid \vx, \vtheta),
    \label{eq:proba_cluster}
\end{equation}
where $\tilde p(\vy \mid \vx, \vtheta)=\tfrac{p(\vy \mid \vx, \vtheta)}{\sum_{i=1}^N p(\vy^{(i)} \mid \vx, \vtheta)}$ for $\vy \in \mathcal{Y}$.
Using the Rao-Blackwellized Monte Carlo estimator, the semantic entropy~\cite{kuhnsemantic,farquhar2024detecting} is approximated as:
\begin{equation}
    H_{\mathrm{SE}}(\mathcal{Y} \mid \vx, \vtheta) = - \sum_{k=1}^K p(\mathcal{C}_k \mid \vx, \vtheta) \log p(\mathcal{C}_k \mid \vx, \vtheta)
    \label{eq:semantic_entropy}
\end{equation}
and measures the dispersion of probability mass across distinct meanings.  
A low $H_{\mathrm{SE}}$ value indicates that model responses are tightly clustered around a single meaning, whereas a high %
value reflects greater semantic diversity in the outputs.

\paragraph{Discrete Semantic Entropy.}
Discrete semantic entropy~\cite{farquhar2024detecting,kuhnsemantic} is another extension of semantic entropy for black-box LVLMs.
It approximates the probability mass $p(\mathcal{C}_k \mid \vx, \vtheta)$ assigned to each cluster $\mathcal{C}_k$ with the empirical cluster probability $\lvert \mathcal{C}_k \rvert/n$, yielding:
\begin{equation}
    H_{\mathrm{DSE}}(\mathcal{Y} \mid \vx, \vtheta) = - \sum_{k=1}^K \tfrac{\lvert \mathcal{C}_k \rvert}{n} \log \tfrac{\lvert \mathcal{C}_k \rvert}{n}. 
    \label{eq:discrete_semantic_entropy}
\end{equation}

\subsection{Latent Uncertainty Estimation}
\label{sec:latent_uq_estimation}
Instead of assessing uncertainty using token likelihoods, alternative approaches leverage dense semantic information retained within the internal states of LVLMs to measure semantic divergence.
\paragraph{Semantic Volume.}
Given a context $\vx$ and a set of generated candidate sequences $\mathcal{Y}:=\{\vy^{(1)}, \ldots, \vy^{(N)}\}$, the semantic volume is defined over the embedding matrix $\mPhi=[\vphi_1, \vphi_2, \ldots, \vphi_N] \in \R^{d \times N}$, where each $\vphi_i \in \R^d$ depicts the sentence embedding of the generated answer $\vy^{(i)}$ within the $d$-dimensional semantic space of the LVLM.
Note that the sentence embedding can be derived either by averaging the token embedding or by taking the last token embedding from a chosen layer or attention head~\cite{azaria2023the, chen2024, janiak2025, ettori2025, orgad2025llms}. 
Let $\mSigma=\mPhi^T\left[\mI_d - \tfrac{\mathbf{1}_d\mathbf{1}_d^T}{d}\right]\mPhi$ denote the empirical covariance matrix of $\mPhi$.
The semantic volume\footnote{The term ``volume'' arises from the geometric interpretation that $\det(\mPhi^T\mPhi)$ is the volume of the parallelepiped spanned by $\{\vphi_i\}_{i=1}^n$.} 
is then defined as~\cite{zhouyin2021, shwartz2023, chen2024}:
\begin{equation}
    V(\mathcal{Y} \mid \vx, \vtheta) = \log \det (\mSigma + \alpha \mI_N) = \sum_{i=1}^N \log(\vlambda_i),
    \label{eq:semantic_volume}
\end{equation}
where $\det(\mX)$ represents the determinant of matrix $\mX$, $\alpha>0$ is a small regularization term to make $\mSigma$ full rank, and $\vlambda=[\vlambda_1, \ldots, \vlambda_N]$ denotes eigenvalues of the matrix $\mSigma + \alpha \mI_N$.
Variants instead define $\mSigma$ using the cosine similarity matrix~\cite{li2025semantic, lau2025uncertainty} or the Gram matrix~\cite{Sriramanan2024} and combine $\mSigma$ with probability responses~\cite{lau2025uncertainty}.
The semantic volume can be interpreted as the differential entropy within the sentence-level LVLM embedding space~\cite{chen2024}.

\section{Semantic Gaussian Process Uncertainty}
\label{sec:Method}

In this section, we present the details of our proposed Semantic Gaussian Process Uncertainty (SGPU) framework, which is designed to quantify semantic uncertainty in both black-box and white-box LVLMs. 
The complete pipeline is depicted in Figure~\ref{fig:pipeline0}.
We start by generating multiple candidate sequences $\mathcal{Y}:=\{\vy^{(1)}, \ldots, \vy^{(N)}\}$ for a given context $\vx$ (Step~1).
Then, using an external sentence embedding model, we embed each answer in a dense semantic space and encode their semantic consistency into
\gianni{a} single vector representation
(Step~2).
Finally, this aggregated representation is passed to a Gaussian Process
\gianni{Classifier}
(GPC), which learns to map patterns of semantic consistency to
\gianni{
the probability that the generated response set is trustworthy
}
(Step~3).
Details and justifications
\gianni{for}
how the semantic consistency of generated answers is encoded are provided in
Section~\ref{sec:SGPU-embeddings}, and the Gaussian Process Classifier is described in
Section~\ref{sec:SGPU-GP}.

\subsection{Semantic Consistency Encoding}
\label{sec:SGPU-embeddings}

\paragraph{Sentence Embedding.}
To capture the semantic consistency of $N$ generated answers $\mathcal{Y}:=\{\vy^{(1)}, \ldots, \vy^{(N)}\}$, we embed each generated answer $\vy^{(i)}$ in $\mathcal{Y}$ in a dense semantic space using an external sentence encoder $E(\cdot)$. 
For each generated answer $\vy^{(i)}$ in $\mathcal{Y}$, we denote its $d$-dimensional normalized embedding vector as $\vphi_i \in \R^d$, with $\lVert \vphi_i \rVert=1$.
The sentence embedding can be obtained either by averaging the token embeddings, by using the embedding of the last token, or by taking the embedding of a designated special token.

This section provides an interpretation of the eigenvalues of the Gram matrix $\mSigma$ of the embedding matrix $\mPhi$ from the answer set $\mathcal{Y}$.  The Gram matrix is %
\begin{equation}
\small
    \mSigma=\mPhi^T\mPhi \in \R^{N \times N}, \text{ where } \mPhi=[\vphi_1, \ldots, \vphi_N] \in \R^{d \times N}.
    \label{def:Sigma}
\end{equation}
\normalsize
Its eigenvalues
\gianni{encode}
information about the semantic consistency among the generated answers in $\mathcal{Y}$.
To formalize it, we introduce the following idealized assumption.

\begin{assumption}[Ideal Sentence Encoding]
    Let $p \geq N$.
    For the set of generated answers $\mathcal{Y}:=\{\vy^{(1)}, \ldots, \vy^{(N)}\}$, we assume there exists an ideal sentence encoder $E^*(\cdot)$ such that
    \begin{equation*}
        \vphi_i^{*T} \vphi_j^* = \left\{
    \begin{array}{ll}
        1 & \mbox{if $\vy^{(i)}$ and $\vy^{(j)}$ share the same meaning} \\
        0 & \mbox{otherwise,}
    \end{array}
\right.
    \end{equation*} 
    where $\vphi_i^* = E^*(\vy^{(i)}) \in \R^p$ depicts the $p$-dimensional normalized embedding vector (with $\lVert \vphi_i^* \rVert=1$) of the generated answer $\vy^{(i)} \in \mathcal{Y}$\footnote{Note that we can have $p \neq d$.}.
    We denote by $\mPhi^*=[\vphi_1^*, \ldots, \vphi_N^*] \in \R^{p \times N}$ the ideal embedding matrix of $\mathcal{Y}$ and by $\mSigma^*=\mPhi^{*T}\mPhi^* \in \R^{N \times N}$ its Gram matrix.
    \label{assumption:ideal_encoder}
\end{assumption}

Lemma~\ref{lemma:ideal_encoder} shows that the eigenvalues of the ideal Gram matrix
$\mSigma^*=\mPhi^{*T}\mPhi^*$
capture meaningful information about the distribution of generated answers across meanings
(see Appendix~\ref{appendix:proof}).

\begin{lemma}[Ideal Semantic Consistency Vector]
     Let $\vlambda^* = [\vlambda_1^*, \dots, \vlambda_N^*]^T \in \R^N$ be the vector of eigenvalues of $\mSigma^*$ ordered such that $\vlambda_1^* \geq \cdots \geq \vlambda_N^* \geq 0$. 
     Then, $\vlambda^*$ depicts the repartition of generated answers $\mathcal{Y}$ across meanings, where:
     \begin{enumerate}
        \item The sum of eigenvalues equals the number of generated answers $N$, i.e., $\sum_{i=1}^N \vlambda_i^*=N$.
         \item The number of nonzero eigenvalues in $\vlambda^*$ corresponds to the number of distinct semantic meanings present in the generated answers $\mathcal{Y}$, i.e., the number of ideal distinct semantic clusters.
         \item Each nonzero $\vlambda_i^*$ is equal to the number of generated answers in $\mathcal{Y}$ that share the same semantic meaning, i.e., the size of an ideal semantic cluster.
     \end{enumerate}
     \label{lemma:ideal_encoder}
\end{lemma}

\begin{remark}
    Intuitively, the vector $\vlambda^*$ serves as a compact descriptor of the semantic consistency within $\mathcal{Y}$.
    High semantic consistency is characterized by a largest eigenvalue
    $\lambda_1^*$ close to $N$, indicating that most or all answers share the same meaning.
    In contrast, a flatter spectrum implies that the generated answers are distributed across several distinct meanings.
\end{remark}

\gianni{Leveraging Weyl's inequality, Lemma~\ref{lemma:semantic_consistency_vector} describes how the eigenvalues of the Gram matrix
$\mSigma=\mPhi^T\mPhi$
derived from an external sentence encoder approximate the ideal semantic consistency vector $\vlambda^*$
(see Appendix~\ref{appendix:proof}).}

\begin{lemma}
     Let $\vlambda \in \R^N$ be the vector of eigenvalues of $\mSigma$ ordered such that $\vlambda_1 \geq \cdots \geq \vlambda_N \geq 0$. 
     For all $i \in [N]$, we have 
     \begin{equation*}
       \vdelta_i = \vlambda_i - \vlambda_i^* \in \bigl[-\sqrt{N(N-1)}\epsilon, \sqrt{N(N-1)}\epsilon\bigr],
    \end{equation*} 
    where $\epsilon=\max_{i,j} \lvert \vphi_i^{T} \vphi_j - \vphi_i^{*T} \vphi_j^* \rvert \in [0, 2]$.
    Besides, $\sum_{i=1}^N \vdelta_i = 0$.
    \label{lemma:semantic_consistency_vector}
\end{lemma}

\begin{remark}
    $\epsilon=\max_{i,j} \lvert \vphi_i^{T} \vphi_j - \vphi_i^{*T} \vphi_j^* \rvert$ measures the maximum semantic similarity error over the generated answers $\mathcal{Y}$ between an ideal sentence encoder $E^*(\cdot)$ and the external sentence encoder $E(\cdot)$.
    A low $\epsilon$ indicates that the semantic consistency vector $\vlambda$ depicts the repartition of generated answers $\mathcal{Y}$ across different meanings (Lemma~\ref{lemma:ideal_encoder}). 
\end{remark}

\begin{remark}
    Although we have $\sum_{i=1}^N \vlambda_i=\sum_{i=1}^N \vlambda_i^*=N$, as the semantic similarity error $\epsilon$ increases, the information captured by $\vlambda$ about the distribution of generated answers $\mathcal{Y}$ across different meanings is perturbed.
    Such perturbation may limit the reliability of direct UQ measures based on the eigenvalues of $\mSigma$, such as the semantic volume~\cite{chen2024} or the Von Neumann entropy~\cite{janiak2025} (see Appendix~\ref{appendix:limitations} for further details on how representation noise affects the semantic consistency vector $\vlambda$).
\end{remark}

We propose to exploit the information about the distribution of generated answers across meanings encoded in the semantic consistency vector $\vlambda$.
To improve robustness to
\gianni{semantic-similarity}
errors, we train a classifier that maps the complete spectral pattern $\vlambda$ to
\gianni{the probability that the generated response set is trustworthy},
as discussed in the following section.

\subsection{Semantic Uncertainty Prediction}
\label{sec:SGPU-GP}

Following prior work~\cite{band2021,kuhnsemantic,farquhar2024detecting,janiak2025,manakul2023},
we formulate the evaluation as a binary classification problem that predicts whether the generated response set can be trusted for a given context. \gianni{
More precisely, SGPU learns the statistical relation between semantic-consistency patterns and response correctness.
It does not assume that consistency alone is sufficient for correctness, nor does it directly verify the factual content of an answer.
}

\paragraph{Binary Classification Problem.}
We assume we have a training dataset $\mathcal{D}:=\{\vx^{(i)}, \mathcal{Y}^{(i)}\}_{i=1}^M$, where $\mathcal{Y}^{(i)}:=\{\vy_i^{(1)}, \ldots, \vy_i^{(N)}\}$ depicts a set of $N$ candidate responses generated by the LVLM for the given context $\vx^{(i)}$.
To frame the problem into a binary classification task, we transform the dataset $\mathcal{D}$ into $\mathcal{\tilde D}:=\{\vlambda^{(i)}, l^{(i)}\}_{i=1}^M$, where $\vlambda^{(i)} \in \R^N$ denotes the semantic consistency vector of the generated responses $\mathcal{Y}^{(i)}$ for the given context $\vx^{(i)}$  (Lemma~\ref{lemma:semantic_consistency_vector}) and $l^{(i)}$ is the binary label indicating the correctness of generated answers $\mathcal{Y}^{(i)}$.
Correctness of predicted answers is commonly used as a proxy for semantic uncertainty~\cite{band2021, kuhnsemantic, farquhar2024detecting, janiak2025, manakul2023} since uncertain generations are less likely to be correct.
Following recommendations in \cite{janiak2025}, for each generated response $\vy_i^{(j)}$ relative to the context $\vx^{(i)}$ in $\mathcal{Y}^{(i)}$, we evaluate the correctness with respect to a reference answer $\bar \vy_i$ using an LLM-as-judge approach. \gianni{
We use different judge models to label the training and test sets.
In particular, the test labels are produced by a larger judge model that is not used to generate the training labels.
This reduces dependence on a single judge and avoids using exactly the same labelling model on both splits.
}
Further details are provided in Appendix~\ref{app_label}.

\paragraph{Gaussian Process Classifier.}
Given a set of generated outputs $\mathcal{Y}:=\{\vy^{(1)}, \ldots, \vy^{(N)}\}$ related to a given context $\vx$, the objective is to predict the truthfulness label $l \in \{0,1\}$ of these generated samples using the semantic consistency vector $\vlambda \in \R^N$ defined in Lemma~\ref{lemma:semantic_consistency_vector}.
We do this by computing the label probability 
\begin{equation}
    p(l \mid \mathcal{Y}, \vx, \vtheta)=p(l \mid \vlambda)
\end{equation}
with a Gaussian Process Classifier (GPC)~\cite{Rasmussen2005} trained on the dataset $\mathcal{\tilde D}:=\{\vlambda^{(i)}, l^{(i)}\}_{i=1}^M$.
In the following, we denote by $\vl=[l^{(i)}, \ldots,l^{(M)}]^T \in \R^M$ and $\mLambda=[\vlambda^{(1)}, \ldots, \vlambda^{(M)}]^T \in \R^{M \times N}$ the collections of labels and semantic consistency vectors, respectively.
For a fixed $\vlambda$, the GPC models $p(l \mid \vlambda)$ as a Bernoulli distribution with parameter $p(l= 1 \mid \vlambda)=s\bigl(f(\vlambda)\bigr)$, where $f(\cdot)$ is a latent function and $s: \R \to [0,1]$ is a sigmoid function.
The latent function $f(\cdot)$ is assumed to follow a Gaussian Process (GP) prior 
\begin{equation}
    f \sim \mathcal{GP}\bigl(m(\vlambda), k(\vlambda, \vlambda')\bigr),
\end{equation}
where $m(\cdot)$ is the mean function and $k(\cdot, \cdot)$ is a kernel function.
This implies that the evaluation vector of $f(\cdot)$ on $\mLambda$ defined as $\vf=[f(\vlambda_1), \ldots, f(\vlambda_M)]^T \in \R^M$ is drawn from a multivariate Gaussian distribution $\vf \sim \mathcal{N}(\vmu, \mK)$, where $\vmu=[m(\vlambda_1), \ldots, m(\vlambda_M)]^T$ and $\mK_{ij}=k(\vlambda_i, \vlambda_j)$.
Since neither of the class labels is considered more probable than the others, the prior mean is usually set to zero. 
The kernel function $k(\cdot, \cdot)$ encodes prior beliefs about the properties of the latent function $f(\cdot)$.
The posterior distribution over the latent values $\vf$ at the observed $\mLambda$ is defined as:
\begin{equation}
    p(\vf \mid \mathcal{\tilde D})=\tfrac{1}{p(\mathcal{\tilde D})} p(\vl \mid \vf)p(\vf \mid \mLambda), \\
\end{equation}
where $p(\mathcal{\tilde D})$ denotes the marginal likelihood and $p(\vl \mid \vf)=\Pi_{j=1}^M s\bigl(l^{(i)}f(\vlambda_i)\bigr)$ the joint likelihood of the independent Bernoulli variables in $\vl$.
The distribution for the latent function value $f=f(\vlambda)$ at a new semantic consistency vector $\vlambda$ is obtained by averaging over the posterior distribution
\begin{equation}
    p(f  \mid \vlambda, \mathcal{\tilde D}) = \int p(f \mid \vf, \mathcal{\tilde D})p(\vf \mid \mathcal{\tilde D})d\vf
\end{equation}
and the predictive label probability $p(l \mid \mathcal{Y}, \vx, \vtheta, \mathcal{\tilde D})$ is computed as the expectation:
\begin{equation*}
    p(l \mid \mathcal{Y}, \vx, \vtheta, \mathcal{D})=p(l \mid \vlambda, \mathcal{\tilde D})=\int p(l \mid f) p(f  \mid \vlambda, \mathcal{\tilde D})df,
\end{equation*}
where $p(l \mid f)=s\bigl(l f(\vlambda)\bigr)$.
Note that neither the marginal likelihood, nor the posterior itself, or predictions can be computed analytically, so approximations such as Monte-Carlo or Laplace approximations are needed~\cite{Rasmussen2005}.
GPC provides the advantage of direct access to its internal uncertainty estimates (through the predictive standard deviation), requires only a small amount of training data, and is typically well-calibrated. 
Moreover, it has strong potential to generalize effectively to new or varying inputs. 
More detailed motivations can be found in Appendix \ref{app_GP_motivation}. 

\section{Experiments}
\label{sec:Experiment}

\subsection{Overall Performance Comparison}

\paragraph{Datasets and Models.}
In this section, we evaluate our method across diverse tasks, datasets, and architectures.
Our evaluation is conducted on visual question answering (VQA), image classification, and
\gianni{text-based question answering (QA)}
tasks.
For VQA, we consider four datasets: \textbf{ADVQA}~\cite{li2021adversarial},
\textbf{VQARAD}~\cite{lau2018dataset},
\textbf{OKVQA}~\cite{marino2019ok}, and
\textbf{VizWiz}~\cite{gurari2018vizwiz}.
For image classification, we use
\gianni{\textbf{CIFAR-10}}
~\cite{krizhevsky2009learning} and
\textbf{Imagenette}~\cite{imagenette};
\gianni{for text-based QA,}
we use \textbf{TriviaQA}~\cite{joshi2017triviaqa} and
\textbf{PopQA}~\cite{chang2023popqa}.
Experiments on vision and VQA tasks are performed with
\texttt{Qwen2.5-VL-3B}~\cite{wei2025deepseek},
\texttt{Qwen2.5-VL-7B}~\cite{bai2025qwen2},
\texttt{llava-mistral-7b}~\cite{liu2024improved},
\gianni{and}
\texttt{idefics2-8b-chatty}~\cite{laurenccon2024matters}.
We conduct experiments with
\texttt{Llama-3.1-8B}~\cite{wei2025deepseek}
for QA datasets.
\gianni{
Dataset splits, generation parameters, training-set sizes, judge models,
random seeds, and implementation details are provided in
Appendix~\ref{app_dataset}.
}

\begin{table}[t]
\centering
\caption{
\gianni{
\textbf{Aggregate comparison of training-free and supervised uncertainty estimators.}
}
Evaluation metrics averaged across all LVLM architectures and datasets.
\textbf{Bold} values
\gianni{indicate}
the best performance, and
\underline{underlined} values indicate the second-best.
}
\label{tab:overall_mean}
\resizebox{0.5\textwidth}{!}{
\begin{tabular}{ccccc}
\toprule
& \textbf{Method}
& \textbf{AUROC$\uparrow$}
& \textbf{AUARC$\uparrow$}
& \textbf{ECE$\downarrow$} \\
\midrule
\multirow{7}{*}{\rotatebox{90}{\gianni{Training-free}}}
& SE & 0.755 & 0.682 & 0.200 \\
& PE & 0.751 & 0.681 & \underline{0.133} \\
& DSE & 0.757 & 0.681 & 0.210 \\
& KLE & 0.765 & 0.685 & 0.206 \\
& UMPIRE & 0.723 & 0.663 & 0.221 \\
& Cov Eigenscore & 0.716 & 0.667 & 0.288 \\
& Cos Eigenscore & 0.746 & 0.685 & 0.161 \\
\midrule
\multirow{4}{*}{\rotatebox{90}{Trained}}
& SAPLMA & 0.761 & 0.687 & 0.200 \\
& SEP & 0.711 & 0.668 & 0.252 \\
& \cellcolor{gray!20} Global-\method~(ours)
& \cellcolor{gray!20} \underline{0.766}
& \cellcolor{gray!20} \underline{0.688}
& \cellcolor{gray!20} 0.155 \\
& \cellcolor{green!20} \textbf{\method} (ours)
& \cellcolor{green!20} \textbf{0.788}
& \cellcolor{green!20} \textbf{0.702}
& \cellcolor{green!20} \textbf{0.115} \\
\bottomrule
\end{tabular}
}
\end{table}

\paragraph{Evaluation Metrics.}
Following previous work~\cite{kuhnsemantic,farquhar2024detecting,janiak2025,lau2025uncertainty},
we evaluate uncertainty methods using AUROC~\cite{hendrycks2016baseline},
AUARC~\cite{kuhnsemantic}, and ECE~\cite{guo2017calibration}.
AUROC measures how well an uncertainty score ranks correct generations above incorrect ones.
Higher scores are better, with a perfect score of $1$ and random ranking corresponding to $0.5$.
AUARC evaluates selective prediction by measuring accuracy as uncertain samples are progressively rejected.
A higher AUARC indicates that errors are concentrated among low-confidence samples.
ECE measures the difference between predicted confidence and empirical correctness, where lower values indicate better calibration. \gianni{
For methods that do not directly output probabilities, ECE is computed after applying the same score-to-confidence mapping fitted only on the validation split.
}
More details on the metrics can be found in Appendix~\ref{app_metrics}.

\paragraph{Baselines.}
We evaluate SGPU using a
\gianni{Matérn}
kernel for the GPC, and generated answers are embedded using
\texttt{all-MiniLM-L6-v2}~\cite{reimers2019sentence}.
We compare our proposal with
popular training-free semantic UQ methods
:
Predictive Entropy (PE)~\cite{kuhnsemantic,farquhar2024detecting},
Semantic Entropy (SE)~\cite{kuhnsemantic},
Discrete Semantic Entropy (DSE)~\cite{farquhar2024detecting},
\joseph{Kernel Language Entropy (KLE) with a Heat Kernel~\cite{nikitin2024kernel}},
UMPIRE~\cite{lau2025uncertainty},
Cov Eigenscore~\cite{chen2024}, and
Cos Eigenscore~\cite{chen2024},
a variant of Cov Eigenscore based on the cosine-similarity matrix
(see Section~\ref{sec:uq_estimation}).
\joseph{
We also compare SGPU with two supervised methods:
SAPLMA~\cite{azaria2023the} and
Semantic Entropy Probes (SEP)~\cite{kossen2024semantic}.
Training procedures, selected layers, hyperparameters, and data splits are detailed in
Appendix~\ref{app_baselines}.
}
\begin{table*}[t]
\centering
\scriptsize
\setlength{\tabcolsep}{2pt}
\caption{
\gianni{
\textbf{Model-wise comparison averaged across the vision datasets.}
}
\textbf{Bold} values
\gianni{indicate}
the best performance, and
\underline{underlined} values indicate the second-best.
}
\label{mean_model}
\resizebox{0.95\textwidth}{!}{
\begin{tabular}{cc|ccc|ccc|ccc|ccc}
\toprule
&& \multicolumn{3}{c}{\texttt{llava-mistral-7b}}
& \multicolumn{3}{c}{\texttt{Qwen2.5-VL-3B}}
& \multicolumn{3}{c}{\texttt{Qwen2.5-VL-7B}}
& \multicolumn{3}{c}{\texttt{idefics2-8b-chatty}} \\
& Method
& AUROC$\uparrow$ & AUARC$\uparrow$ & ECE$\downarrow$
& AUROC$\uparrow$ & AUARC$\uparrow$ & ECE$\downarrow$
& AUROC$\uparrow$ & AUARC$\uparrow$ & ECE$\downarrow$
& AUROC$\uparrow$ & AUARC$\uparrow$ & ECE$\downarrow$ \\
\midrule
\multirow{7}{*}{\rotatebox{90}{\gianni{Training-free}}}
& SE
& 0.780 & 0.657 & 0.191
& 0.769 & 0.681 & 0.143
& 0.744 & 0.726 & 0.271
& 0.725 & 0.665 & 0.193 \\

& PE
& 0.781 & 0.657 & 0.152
& 0.777 & 0.685 & \underline{0.103}
& 0.745 & 0.729 & \textbf{0.109}
& 0.700 & 0.653 & \underline{0.172} \\

& DSE
& 0.782 & 0.656 & 0.192
& 0.776 & 0.682 & 0.156
& 0.744 & 0.723 & 0.271
& 0.725 & 0.663 & 0.220 \\

& KLE
& 0.750 & 0.645 & 0.225
& \underline{0.787} & \underline{0.694} & 0.178
& \underline{0.766} & \underline{0.739} & 0.186
& \textbf{0.755} & 0.661 & 0.235 \\

& UMPIRE
& 0.773 & 0.654 & 0.191
& 0.714 & 0.649 & 0.159
& 0.704 & 0.703 & 0.310
& 0.703 & 0.645 & 0.223 \\

& Cov Eigenscore
& 0.729 & 0.648 & 0.198
& 0.679 & 0.629 & 0.294
& 0.741 & 0.730 & 0.339
& 0.713 & 0.660 & 0.321 \\

& Cos Eigenscore
& 0.758 & 0.670 & 0.180
& 0.737 & 0.667 & 0.112
& 0.743 & 0.726 & 0.174
& 0.746 & \underline{0.678} & 0.178 \\

\midrule
\multirow{4}{*}{\rotatebox{90}{Trained}}
& SAPLMA
& 0.787 & 0.655 & \underline{0.129}
& 0.763 & 0.672 & 0.175
& 0.748 & 0.730 & 0.183
& \underline{0.747} & \textbf{0.692} & 0.313 \\

& SEP
& 0.758 & \underline{0.681} & 0.219
& 0.699 & 0.633 & 0.237
& 0.682 & 0.699 & 0.252
& 0.705 & 0.661 & 0.301 \\

& \cellcolor{gray!20} Global-\method~(ours)
& \cellcolor{gray!20} \underline{0.799}
& \cellcolor{gray!20} \underline{0.681}
& \cellcolor{gray!20} 0.159
& \cellcolor{gray!20} 0.786
& \cellcolor{gray!20} 0.689
& \cellcolor{gray!20} 0.127
& \cellcolor{gray!20} 0.755
& \cellcolor{gray!20} 0.731
& \cellcolor{gray!20} 0.135
& \cellcolor{gray!20} 0.725
& \cellcolor{gray!20} 0.661
& \cellcolor{gray!20} 0.198 \\

& \cellcolor{green!20} \textbf{\method} (ours)
& \cellcolor{green!20} \textbf{0.819}
& \cellcolor{green!20} \textbf{0.685}
& \cellcolor{green!20} \textbf{0.108}
& \cellcolor{green!20} \textbf{0.806}
& \cellcolor{green!20} \textbf{0.704}
& \cellcolor{green!20} \textbf{0.071}
& \cellcolor{green!20} \textbf{0.782}
& \cellcolor{green!20} \textbf{0.748}
& \cellcolor{green!20} \underline{0.115}
& \cellcolor{green!20} 0.743
& \cellcolor{green!20} 0.669
& \cellcolor{green!20} \textbf{0.164} \\
\bottomrule
\end{tabular}
}
\end{table*}

\paragraph{Results and Discussion.}
Tables~\ref{tab:overall_mean} and~\ref{mean_model}
present aggregate and model-wise comparisons across the evaluated datasets and architectures.
\gianni{
SGPU achieves the strongest aggregate AUROC, AUARC, and ECE in
Table~\ref{tab:overall_mean}.
At the model level, SGPU obtains the best AUROC and AUARC for three of the four LVLMs and the best ECE for three of the four LVLMs.
We therefore describe SGPU as providing strong and often state-of-the-art performance rather than uniformly outperforming every baseline.
} Due to the open-ended nature of free-form generation, uncertainty estimation is generally more challenging for VQA than for image classification.
\gianni{
Dataset-level results for VQARAD, ADVQA, and the remaining benchmarks are reported in Appendix~\ref{app_additional}.
Confidence intervals and paired comparisons should be used before describing differences as statistically significant.
}

\paragraph{Detection of Unsafe SGPU Predictions.}
As shown in Figure~\ref{app_fig:gp_histo}, one motivation for using a GPC is its ability to estimate the predictive variance
\gianni{of each SGPU prediction}.
This variance reflects the confidence of SGPU
\gianni{in its own prediction and should not be interpreted as the internal uncertainty of the LVLM}.
In practice, this allows us to flag cases in which SGPU is not confident enough to make a reliable judgment.
\gianni{
The exact rejection criterion, its coverage, and accuracy before and after rejection are reported in Appendix~\ref{app_gp_internal_uncertainty}.
}

\subsection{Transfer Experiments and Generalization}
\label{sec:transfer}

\gianni{While SGPU is trained and evaluated within a specific model and dataset setting, Global-SGPU pools training data from multiple LVLMs and datasets to support transfer to unseen architectures, datasets, and modalities.}
\joseph{
SGPU attains its strongest performance when trained and evaluated on the same architecture, but its external spectral representation also supports transfer across models.
Global-\method~is trained by pooling response sets from several source models and datasets.
For each zero-shot transfer experiment, the target model and target dataset must be excluded from training and hyperparameter selection; the exact source--target splits are reported in Appendix~\ref{app_transfer}.
Unlike SAPLMA and SEP, which depend on model-specific internal representations and therefore require retraining or feature adaptation, Global-\method~can be directly applied to response sets generated by an unseen architecture.
On the cross-modality evaluation in Table~\ref{tab:transfer_trivia_qa},
Global-\method~achieves the best mean AUROC and remains competitive in AUARC and ECE, although it does not outperform every baseline on every metric.
}

\joseph{
\subsection{\gianni{Mitigating Failures of Self-Consistency}}
}

\joseph{
Direct self-consistency methods tend to assign high confidence when a model repeatedly produces the same answer, even when that answer is incorrect.
SGPU can identify some of these failures.
On \texttt{Qwen2.5-VL-7B}/\textbf{CIFAR}, among the 4{,}058 response sets in which the same answer appeared at least ten times, SGPU assigned low confidence to 304 response sets that were indeed incorrect.
This shows that SGPU mitigates, rather than completely resolves, the failure of direct self-consistency.
A complete confusion matrix, precision, recall, and qualitative examples are provided in Appendix~\ref{app_self_consistency}.
}

\begin{table}[t]
\centering
\caption{
\joseph{
\gianni{
\textbf{Cross-modality transfer of Global-\method.}
Mean results on unseen combinations of
\texttt{Qwen2.5-3B-Instruct} and \texttt{Llama-3.1-8B}
with \textbf{TriviaQA} and \textbf{GSM8K}.
The target models and datasets are excluded from Global-\method~training.
}
}
}
\label{tab:transfer_trivia_qa}

\resizebox{0.35\textwidth}{!}{
\begin{tabular}{cccc}
\toprule
\textbf{Method}
& \textbf{AUROC$\uparrow$}
& \textbf{AUARC$\uparrow$}
& \textbf{ECE$\downarrow$} \\
\midrule
SE & 0.752 & 0.452 & 0.286 \\
DSE & 0.759 & \textbf{0.503} & \textbf{0.265} \\
KLE & \underline{0.795} & 0.471 & 0.348 \\
\rowcolor{gray!20}
Global-\method~(ours)
& \textbf{0.796}
& \underline{0.475}
& \underline{0.268} \\
\bottomrule
\end{tabular}
}
\end{table}

\subsection{Computational Cost}

We report the inference-time comparison of semantic UQ methods in
Table~\ref{tab:time_compute}.
Since the same sampled answers are reused by all methods, we first compare only the
\gianni{post-hoc}
cost required to compute the uncertainty estimate.
\gianni{
For SGPU, this cost includes sentence encoding, construction of the Gram matrix,
eigendecomposition, and GPC prediction, but excludes answer generation.
The comparison therefore isolates the uncertainty-estimation overhead and should not be interpreted as the complete deployment cost.
}
As shown in Table~\ref{tab:time_compute}, SGPU has a lower post-hoc overhead than KLE and a cost comparable to SE.

\begin{table}[h!]
\centering
\caption{
\gianni{
Average post-hoc inference time over 100 inputs.
Response generation is excluded.
}
}
\label{tab:time_compute}
\resizebox{0.6\columnwidth}{!}{
\begin{tabular}{ccc}
\hline
\textbf{Method} & \textbf{OKVQA} & \textbf{VizWiz} \\
\hline
\rowcolor{green!20}
\textbf{\method} (ours) & 0.66 s & 1.01 s \\
KLE & 2.01 s & 2.61 s \\
SE & 0.72 s & 1.04 s \\
\hline
\end{tabular}
}
\end{table}

\section{Conclusion}

Measuring semantic uncertainty in LVLMs is of critical importance for improving their reliabilty.
In this work, we propose Semantic Gaussian Process Uncertainty (SGPU), a new Bayesian framework for quantifying semantic uncertainty in LVLMs that avoids explicit clustering and operates on the eigenspectrum of answer embeddings.
Our approach provides three main advantages. 
First, \method\ achieves strong and often state-of-the-art performance across models and datasets on standard uncertainty-evaluation metrics such as AUROC, AUARC, and ECE, with particularly strong results on AUROC. 
Second, \method~ works in a fully black-box setting: it does not require access to internal model features, embeddings, or token-level probabilities. This makes it applicable to most real-world LVLMs where internal access is not available. 
Third, by using a Gaussian Process Classifier, \method~ can quantify its own uncertainty while predicting the uncertainty of the LVLM. 
This extra layer of confidence estimation makes the final predictions more stable and reliable. Overall, \method~ offers a simple, scalable, and robust way to measure uncertainty in modern LVLMs, in both white-box and black-box scenarios.

\bibliography{aaai2027}

\clearpage
\clearpage

\setcounter{page}{1}
\appendix
\renewcommand{\thetable}{A.\arabic{table}}
\renewcommand{\thefigure}{A.\arabic{figure}}

\begin{center}
  \Large \textbf{Improving Semantic Uncertainty Quantification in LVLMs with Semantic Gaussian Processes\\ --Supplementary Material--}
\end{center}

\vspace{0.8cm}

\section{Proofs of Section~\ref{sec:SGPU-embeddings}}
\label{appendix:proof}
This section provides the proofs of Lemma~\ref{lemma:ideal_encoder} and Lemma~\ref{lemma:semantic_consistency_vector} discussed in Section~\ref{sec:SGPU-embeddings}.
We start by introducing the following lemma.
\begin{lemma}
    Let $\mPhi^*=[\vphi_1^*, \ldots, \vphi_N^*] \in \R^{p \times N}$ denote the embedding matrix of an ideal sentence encoder $E^*(\cdot)$ defined as in Assumption~\ref{assumption:ideal_encoder}.
    Let $\mathcal{E}:=\{\ve_1, \ldots, \ve_p\}$ the standard basis of $\R^p$.
    Then, there exists an orthogonal matrix $\mP \in \R^{p \times p}$ such that 
    \begin{equation}
        \mPhi^* = \mP \mU,
    \end{equation}
    where
    $
        \mU = \begin{bmatrix}
            \vu_1 \, \vu_2 \, \ldots \ \vu_N \end{bmatrix} \in \R^{p \times N} 
    $ with $\vu_i \in \mathcal{E}$ for all $i$. 
    \label{lem:decomposition_Phi_star}
\end{lemma}
\begin{proof}
    Under Assumption~\ref{assumption:ideal_encoder}, there exists a unique set $\mathcal{V}:=\{\vv_1, \ldots, \vv_k\} \in \{\vphi_1^*, \ldots, \vphi_N^*\}$ of orthonormal vectors in $\R^p$.
    The set $\mathcal{V}$ can be completed with $p-k$ additional orthonormal vectors to form a basis $\mathcal{B}$ of $\R^p$.
    Let $\mP \in \R^{p \times p}$ denote the matrix whose columns are the vectors $\mathcal{B}$.
    $\mP$ is an orthonormal matrix and depicts the change of basis matrix from $\mathcal{B}$ to the standard basis $\mathcal{E}$ of $\R^p$ ($\mP=\mP\mI_p$).
    We obtain $\mU=\mP^T\mPhi^*$, which concludes the proof.
\end{proof}
Leveraging Lemma~\ref{lem:decomposition_Phi_star} and the Weinstein–Aronszajn identity, the following Lemma~\ref{lemma:ideal_encoder} shows that the eigenvalues of the Gram matrix $\mSigma^*=\mPhi^{*T}\mPhi^* \in \R^{N \times N}$ obtained from an ideal sentence encoder capture meaningful information about the semantic consistency among the generated answers in $\mathcal{Y}$ across meaning.
\renewcommand{\thelemma}{1} 
\begin{lemma}[Ideal Semantic Consistency Vector]
     Let $\vlambda^* = [\vlambda_1^*, \dots, \vlambda_N^*]^T \in \R^N$ be the vector of eigenvalues of $\mSigma^*=\mPhi^{*T}\mPhi^*$ ordered such that $\vlambda_1^* \geq \cdots \geq \vlambda_N^* \geq 0$. 
     Then, $\vlambda^*$ depicts the repartition of generated answers $\mathcal{Y}$ across meanings, where:
     \begin{enumerate}
        \item The sum of eigenvalues equals the number of generated answers $N$, i.e., $\sum_{i=1}^N \vlambda_i^*=N$.
         \item The number of nonzero eigenvalues in $\vlambda^*$ corresponds to the number of distinct semantic meanings present in the generated answers $\mathcal{Y}$, i.e., the number of ideal distinct semantic clusters.
         \item Each nonzero $\vlambda_i^*$ is equal to the number of generated answers in $\mathcal{Y}$ that share a same semantic meaning, i.e., the size of an ideal semantic cluster.
     \end{enumerate}
\end{lemma}
\begin{proof}
    Let $\mathcal{E}:=\{\ve_1, \ldots, \ve_p\}$ the standard basis of $\R^p$.
    From Lemma~\ref{lem:decomposition_Phi_star}, there exists an orthogonal matrix $\mP \in \R^{p \times p}$ such that $$
        \mPhi^*=\mP\mU,
    $$
    where $
        \mU = \begin{bmatrix}
            \vu_1 \, \vu_2 \, \ldots \ \vu_N \end{bmatrix} \in \R^{p \times N} 
    $ with $\vu_i \in \mathcal{E}$ for all $i$.
    From the Weinstein–Aronszajn identity, the matrices 
    $$
        \mSigma^*=\mPhi^{*T} \mPhi^* = \mU^T \underbrace{\mP^T \mP}_{= \mI_p} \mU = \mU^T \mU \in \R^{N \times N}
    $$
    and 
    $$
        \mD=\mU \mU^T \in \R^{p \times p}
    $$ 
    share the same nonzero eigenvalues. 
    Let $i, j \in [p]$.
    We have 
    \begin{align*}
        \mD_{ij}=\bigl[\mU \mU^T \bigr]_{ij} = \sum_{k=1}^N \mU_{ik} \mU_{jk} = \sum_{k=1}^N \bigl[\vu_k \bigr]_i \bigl[\vu_k \bigr]_j.
    \end{align*}
    For all $k$, $\vu_k \in \mathcal{E}$ is a vector in the standard base of $\R^p$, i.e., a vector which has a single element equal to $1$ and all other components equal to $0$.
    Therefore, for all $k$ and $i \neq j$, we have $\bigl[\vu_k \bigr]_i \bigl[\vu_k \bigr]_j=0$.
    For all $i \neq j$, we deduce that $\mD_{ij}=\bigl[\mU \mU^T \bigr]_{ij}=0$ and that $\mD$ is a diagonal matrix.
    Nonzero eigenvalues of $\mSigma^*$ are thus diagonal elements of $\mD$.
    Furthermore, for all $i$, we have 
    \begin{align*}
        \mD_{ii}= \sum_{k=1}^N \bigl[\vu_k \bigr]_i^2&= \sum_{k=1}^N \bigl[\vu_k \bigr]_i=c_i,
    \end{align*}
    where $c_i$ depicts the number of occurences of $\ve_i$ in $\mU$.
\end{proof}
Leveraging the Weyl's inequality, Lemma~\ref{lemma:semantic_consistency_vector} depicts how the eigenvalues of the Gram matrix $\mSigma=\mPhi^{T}\mPhi$ (\eqref{def:Sigma}) derived from an external sentence encoder approximate the ideal semantic consistency vector $\vlambda^*$.
\renewcommand{\thelemma}{2} 
\begin{lemma}
     Let $\vlambda \in \R^N$ be the vector of eigenvalues of $\mSigma$ ordered such that $\vlambda_1 \geq \cdots \geq \vlambda_N \geq 0$. 
     For all $i \in [N]$, we have 
     \begin{equation*}
       \vdelta_i = \vlambda_i - \vlambda_i^* \in \bigl[-\sqrt{N(N-1)}\epsilon, \sqrt{N(N-1)}\epsilon\bigr],
    \end{equation*} 
    where $\epsilon=\max_{i,j} \lvert \vphi_i^{T} \vphi_j - \vphi_i^{*T} \vphi_j^* \rvert \in [0, 2]$.
    Besides, $\sum_{i=1}^N \vdelta_i = 0$.
\end{lemma}
\begin{proof}
    Let $\mDelta=\mSigma-\mSigma^* \in \R^{N \times N}$ and $\epsilon=\max_{i,j} \lvert \mDelta_{ij} \rvert$. 
    Note that $\mDelta$ is symmetric and that $\Tr(\mDelta)=0$.
    We can observe that 
    \begin{equation*}
        \mSigma = \mSigma^* + \bigl[\mSigma-\mSigma^*\bigr] = \mSigma^* + \mDelta.
    \end{equation*}
    From equation above and the Weyl's inequality, for all $i \in [N]$, we have 
    \begin{equation*}
         \lvert \vdelta_i \rvert = \lvert \vlambda_i - \vlambda_i^* \rvert \leq \max \bigl(\lvert \vlambda_N(\mDelta) \rvert, \vlambda_1(\mDelta) \bigr) = \lVert \mDelta \rVert_2
    \end{equation*}
    where $\vlambda_N(\mDelta) \in \R$ and $\vlambda_1(\mDelta) > 0$ denote the minimum et maximum eigenvalues of $\mDelta$.
    From the equivalence norm, we have
    \begin{equation*}
        \lVert \mDelta \rVert_2 \leq \lVert \mDelta \rVert_F,
    \end{equation*}
    where $\lVert \cdot \rVert_F$ is the Frobenius norm.
    By construction of $\mSigma$ and $\mSigma^*$, all diagonal elements of $\mDelta=\mSigma-\mSigma^*$ are equal to zeros.
    We obtain thus:
    \(begin{equation*})
        \lVert \mDelta \rVert_F = \sqrt{\sum_{i=1}^N \sum_{j=1}^N \mDelta_{ij}^2} = \sqrt{\sum_{i=1}^N \sum_{j \neq i}\mDelta_{ij}^2} \leq \sqrt{N(N-1)\epsilon^2} = \sqrt{N(N-1)}\epsilon.
    \)
    Finally, we have $\sum_{i=1}^N \vdelta_i=\sum_{i=1}^N \vlambda_i - \vlambda_i^* = \Tr(\mSigma) - \Tr(\mSigma^*)=N-N=0$.
\end{proof}

\section{Impacts of Embeddings on the Semantic Consistency}
\label{appendix:limitations}

\subsection{Semantic Similarities in UQ}

As mentioned in Section~\ref{sec:background}, semantic UQ methods estimate uncertainty by computing semantic similarities between generated answers either to perform semantic clustering or to directly derive a semantic UQ score, e.g., the semantic volume. 
These methods can rely on two types of semantic spaces for such comparisons.  

\paragraph{Natural Language Inference (NLI).}
In practice, semantic similarities are typically obtained using predictions from NLI models. 
However, despite their popularity, NLI models often struggle to capture the full range of semantic properties in text and are highly sensitive to minor variations in wording, additional correct information, or the presence of non-essential words in generated responses~\cite{arakelyan2024, grewal2024improving} as illustrated in Table~\ref{tab:semantic_equivalence}.

\paragraph{Semantic Embeddings.}
Another line of work explores computing semantic similarities from embeddings derived from the hidden states of LVLMs~\cite{chen2024,binkowski2025, lee2025, janiak2025}. 
Yet, such representations are not specifically designed for this purpose, and these methods are limited to white-box access to the underlying LLM.
Following prior works~\cite{grewal2024improving, abdaljalil2025}, we propose instead to leverage external sentence encoders that are specifically trained to derive semantically meaningful sentence~\cite{reimers2019sentence} and are used in applications such as retrieval~\cite{lewis2020}, classification~\cite{reimers2019sentence} or clustering~\cite{abdaljalil2025}.
In the following section, we indeed observe that embeddings of generated answers derived from an external sentence encoder capture the semantic consistency among those answers more effectively than hidden representations of LVLMs. 
\begin{table}[t]
\centering
\caption{Comparison of bidirectional entailment and cosine similarity for assessing semantic equivalence~\cite{grewal2024improving}. 
Bidirectional entailment scores are obtained using DeBERTaLarge~\cite{he2021deberta} NLI model, while cosine similarity is computed with sentence-BERT~\cite{reimers2019sentence}.}
\label{tab:semantic_equivalence}
\resizebox{1.0\columnwidth}{!}
{
    \begin{tabular}{p{4cm}p{7cm}cc}
        \toprule    
        \textbf{Context} & \textbf{Responses} & \textbf{Bidirectional  Entailment} & \textbf{Cosine Similarity} \\
        \midrule
        
        What is the primary function of the mitochondria in cells? 
        &
        1. The mitochondria produce energy for the cells.\newline
        2. Mitochondria provides energy to cells in the body.
        &
        False & 0.974 \\
        
        \midrule
        
        What happens when you heat ice?
        &
        1. Heating ice will eventually boil after becoming water.\newline
        2. When ice is heated, it melts into water before boiling.
        &
        False & 0.893 \\
        
        \midrule
        
        What do mammals have in common?
        &
        1. Mammals are warm-blooded and have hair or fur.\newline
        2. All mammals (like humans and dogs) are warm-blooded creatures with hair.
        &
        False & 0.927 \\
        
        \bottomrule
    \end{tabular}
}
\end{table}
\newpage
\subsection{Impacts of Embeddings on the Semantic Consistency Vector}
In Section~\ref{sec:SGPU-embeddings} of the main paper, we introduce the semantic consistency vector $\vlambda \in \R^N$ defined as the eigenvalues of the Gram matrix computed from the embeddings of generated answers.
This vector provides a compact yet rich descriptor of semantic consistency of generated answers.
In particular, Lemma~\ref{lemma:ideal_encoder} interprets the eigenvalues $\vlambda^* \in \R^N$ of the Gram matrix derived from ideal embeddings of generated answers (Assumption~\ref{assumption:ideal_encoder}) as the repartition of generated answers across their meanings.
In practice, however, the embeddings are obtained from hidden representations or external models, yielding only an approximation of the Gram matrix associated with an ideal encoder. 
As highlighted by Lemma~\ref{lemma:semantic_consistency_vector}, such approximations can perturb the information captured by the eigenvalues $\vlambda \in \R^N$ regarding the distribution of generated answers, thereby limiting the direct use of these values in UQ methods, e.g., by computing the semantic volume~\cite{chen2024} or the von Neumann entropy~\cite{janiak2025}.

\begin{table}[t]
    \caption{\textbf{Impacts of Embeddings on the Semantic Consistency Vector} over $100$ samples from the \textbf{OKVQA} dataset. For each sample we prompt \texttt{Qwen2.5-VL-7B} to generate $10$ paraphrases of the reference answer and $5$ paraphrases of a semantically distinct answer to get two clusters of $10$ and $5$ responses. Report of the average of the two largest eigenvalues $\vlambda_1$ and $\vlambda_2$ of the Gram matrices computed for each embedding type across all samples and of the maximum semantic similarity error $\epsilon$ (Lemma~\ref{lemma:semantic_consistency_vector}).}
    \label{tab:eigenval2}
    \centering
    \begin{tabular}{c | c c | c}
        \toprule
        \textbf{Sentence Embedding}      & $\vlambda_1$ & $\vlambda_2$ & $\epsilon$ \\
        \midrule
        Ideal Embedding & 10.0 & 5.00 & 0.00 \\
        External Embedding & ~8.41~ & ~3.96~ & ~0.49~\\
        Internal Embedding (layer 16)~   & 11.73 & 1.85 & 0.67\\
        \midrule
    \end{tabular}
\end{table}

\paragraph{Impact of Embeddings.}
To investigate the impact of embeddings on the information captured by the semantic consistency vector $\vlambda = [\vlambda_1 \ldots \vlambda_N ]^T \in \R^N$ (defined in Lemma~\ref{lemma:semantic_consistency_vector}), we conduct a pilot study using $100$ samples from the \textbf{OKVQA} dataset.
In particular, for each sample, we prompt \texttt{Qwen2.5-VL-7B} to generate $10$ paraphrases of the reference answer and $5$ paraphrases of a semantically distinct answer to get two clusters of $10$ and $5$ responses, respectively.
We evaluate two kinds of sentence embeddings: one derived from the hidden representation of the LVLM at an intermediate layer\footnote{We choose the intermediate layer that provides the best approximation of the ideal Gram matrix.}, and another from the external sentence encoder \texttt{all-MiniLM-L6-v2} used in the main experiments.
Table~\ref{tab:eigenval2} reports the average of the two largest eigenvalues of the Gram matrices computed for each embedding type across all samples.
According to Lemma~\ref{lemma:ideal_encoder}, the eigenspectrum of the Gram matrix obtained from an ideal sentence encoder should reflect the distribution of sentences across their meanings.
In this pilot study, the expected eigenspectrum should contain two nonzero eigenvalues equal to $10$ and $5$.
Our results indicate that semantic consistency vectors derived from \texttt{all-MiniLM-L6-v2} capture information about the distribution of generated answers across different meanings more effectively than those based on hidden representations.
Although semantic consistency vectors derived from external encoders provide a more accurate representation of the underlying semantic distribution, the information they convey remains perturbed, which limits their direct applicability for computing a UQ score.
These findings motivate the introduction of SGPU, which more effectively leverages the semantic consistency information encoded in the eigenspectrum despite the approximation errors introduced by the external sentence encoder.

\section{Experimental Settings}

In this section, we provide additional details about the experimental protocol used in Section~\ref{sec:Experiment}.

\subsection{Datasets}
\label{app_dataset}
We evaluate SGPU on Visual Question Answering (VQA), image classification, and textual Question Answering (QA) tasks. 

\paragraph{Visual Question-Answering.}
For VQA, we consider four datasets: 
\textbf{ADVQA}~\cite{li2021adversarial}, which consists of a training set of $6,000$ samples and a test set of $2,000$ samples both drawn from the original training set;
\textbf{VQARAD}~\cite{lau2018dataset}, which consists of a training set of $2,000$ samples from the original training set and a test set of $500$ samples from the original test set;
\textbf{OKVQA}~\cite{marino2019ok}, which consists of a training set of $9,000$ samples drawn from the original training set and a test set of $5,000$ samples drawn from the original validation set;
and \textbf{VizWiz}~\cite{gurari2018vizwiz}, which consists of a training set of $10,000$ samples and a test set of $2,000$ samples both drawn from the original training set.

\paragraph{Image Classification.}
For image classification, we use the \textbf{CIFAR-10}~\cite{krizhevsky2009learning} and \textbf{Imagenette}~\cite{deng2009imagenet} datasets, each split into a training set of $10,000$ samples drawn from the original training data and a test set of $5,000$ samples taken from the original validation set.

\paragraph{Question-Answering.}
For text-based QA, we use the \textbf{TriviaQA}~\cite{joshi2017triviaqa} and \textbf{PopQA}~\cite{chang2023popqa} datasets, each split into a training set of $10,000$ samples drawn from the original training data and a test set of $1,000$ samples taken from the original test set.

\subsection{Baselines}
\label{app_baselines}

We compare SGPU with popular uncertainty-based methods using both the logits and hidden representations of LVLMs.

\paragraph{Logit-based Methods.}
For logit-based measures, we consider the following: \textbf{Predictive Entropy (PE)}~\cite{farquhar2024detecting, lin2024generating, kadavath2022}; 
\textbf{Semantic Entropy (SE)}~\cite{kuhnsemantic, farquhar2024detecting} defined in \eqref{eq:semantic_entropy};
\textbf{Discrete Semantic Entropy (DSE)}~\cite{kuhnsemantic, farquhar2024detecting} defined in \eqref{eq:discrete_semantic_entropy};
\textbf{Kernel Language Entropy (KLE)}~\cite{nikitin2024kernel}, which captures semantic similarities by applying a distance measure in the space of the generated answers.
In particular, KLE uses an external NLI model to construct a semantic graph over the generated responses, and then computes the von Neumann entropy of a graph kernel formed by the combination of the graph Laplacian with a predefined kernel.

\paragraph{Latent-based Methods.}
We also examine semantic UQ methods that exploit dense semantic information retained within the internal states of LVLMs.
In particular, we consider:
\textbf{Cos Eigenscore}~\cite{li2025semantic, lau2025uncertainty} and \textbf{Cov Eigenscore}~\cite{chen2024}, which are semantic volume-based measures (see \eqref{eq:semantic_volume}).
These methods compute the semantic volume using the cosine similarity matrix and the empirical covariance matrix of the feature representations, respectively.
We also include \textbf{UMPIRE}~\cite{lau2025uncertainty}, which combines the semantic volume with the probabilities of generated answers.

\subsection{Metrics}
\label{app_metrics}
\paragraph{AUROC (Area Under the ROC Curve).}
We assign a probability score to each example for belonging to the positive class (positive reflects that the LVLM is certain about its answer). AUROC~\cite{hendrycks2016baseline} measures how well these scores rank true positives above true negatives. It is the chance that a randomly chosen positive example receives a higher score than a randomly chosen negative one. 
Higher AUROC means better class separation.

\paragraph{AUARC (Area Under Accuracy–Retention Curve).}
Each prediction has an associated uncertainty. 
We sort predictions from most confident to least confident and progressively discard (abstain on) the most uncertain ones. 
For every retention level (fraction kept), we compute the accuracy on the retained subset. 
AUARC~\cite{kuhnsemantic} summarizes how accuracy improves as we keep only confident predictions. 
Higher AUARC means uncertainty is useful for selective prediction.

\paragraph{ECE (Expected Calibration Error).}
Each prediction has a confidence value. 
We group predictions into bins of similar confidence and compare, within each bin, the average confidence to the actual fraction that is correct. 
ECE~\cite{guo2017calibration} is the average mismatch across bins. 
Lower ECE means the reported confidences align well with true correctness (the model is better calibrated).

\subsection{Sensitivity to Design Choices}

\paragraph{Sensitivity to Kernel Choices.} For Global-\method~ we swept a set of standard kernels (RBF and Matern with $\nu \in \{0.5, 1.5, 2.5\}$) on the merged training set, and found the resulting inner-validation AUROC to vary only marginally across configurations (Table~\ref{tab:kernel_sweep}), indicating that our method is robust to the choice of kernel. The Matern kernel with $\nu=1.5$ achieved the best inner-validation AUROC and was retained for all subsequent experiments.

\begin{table}[h]
\centering
\caption{Inner-validation AUROC of Global-\method~ for kernel configurations on the merged training set.}
\label{tab:kernel_sweep}
\begin{tabular}{lc}
\toprule
Kernel & Inner-val AUROC \\
\midrule
RBF & 0.748 \\
Matern, $\nu=0.5$ & 0.749 \\
Matern, $\nu=1.5$ & \textbf{0.750} \\
Matern, $\nu=2.5$ & 0.750 \\
\bottomrule
\end{tabular}
\end{table}

\paragraph{Sensitivity to External Encoder Choices.}
Similarly, we examine the impact of the choice of the external sentence encoder used by SGPU. 
Experiments depicted in Table~\ref{tab:ext_enco_HP_SGPU} show that SGPU remains robust to the choice of external sentence embedding. 
In our implementation, SGPU uses the \texttt{all-MiniLM-L6-v2} encoder, which provides a good trade-off between efficiency and representation quality. 

\begin{table}[t]
\caption{\textbf{SGPU is robust to the choice of the external sentence encoder}. Sensitivity analysis of the impact of different external sentence encoders on SGPU performance on the \textbf{VizWiz} dataset using \texttt{Qwen2.5-VL-7B}.}
\label{tab:ext_enco_HP_SGPU}
\centering
\resizebox{0.5\textwidth}{!}{
\begin{tabular}{c|c c c}
\toprule
\textbf{External Encoder}                        & \textbf{AUROC$\uparrow$} & \textbf{AUARC$\uparrow$} & \textbf{ECE$\downarrow$}  \\
\midrule
 \texttt{all-MiniLM-L6-v2}                 & 0.821 & 0.797 & 0.061 \\
 \texttt{all-MiniLM-L12-v2}                & 0.815 & 0.792 & 0.068 \\
 \texttt{modernbert-embed-large}                  & 0.810 & 0.770 & 0.072 \\

\midrule
\end{tabular}
}
\end{table}

\subsection{Other Experimental Settings}
\label{app_exp_setting}
For our experiments, we use the Transformer library~\cite{wolf2020}. 
For each sample, we sample $20$ candidate answers using \texttt{top\_p=0.9}, \texttt{top\_k=50}, and \texttt{temperature=1.0}. 
All models perform inference in \texttt{torch.float16} precision on NVIDIA H100 NVL GPUs. 
We use \texttt{all-MiniLM-L6-v2}~\cite{reimers2019sentence} for sentence embedding with mean pooling and use the Deberta-large for NLI~\cite{he2021deberta}.

\section{Truthfulness Labeling}
\label{app_label}
\begin{figure*}[t!]
    \centering
    \includegraphics[width=1\linewidth]{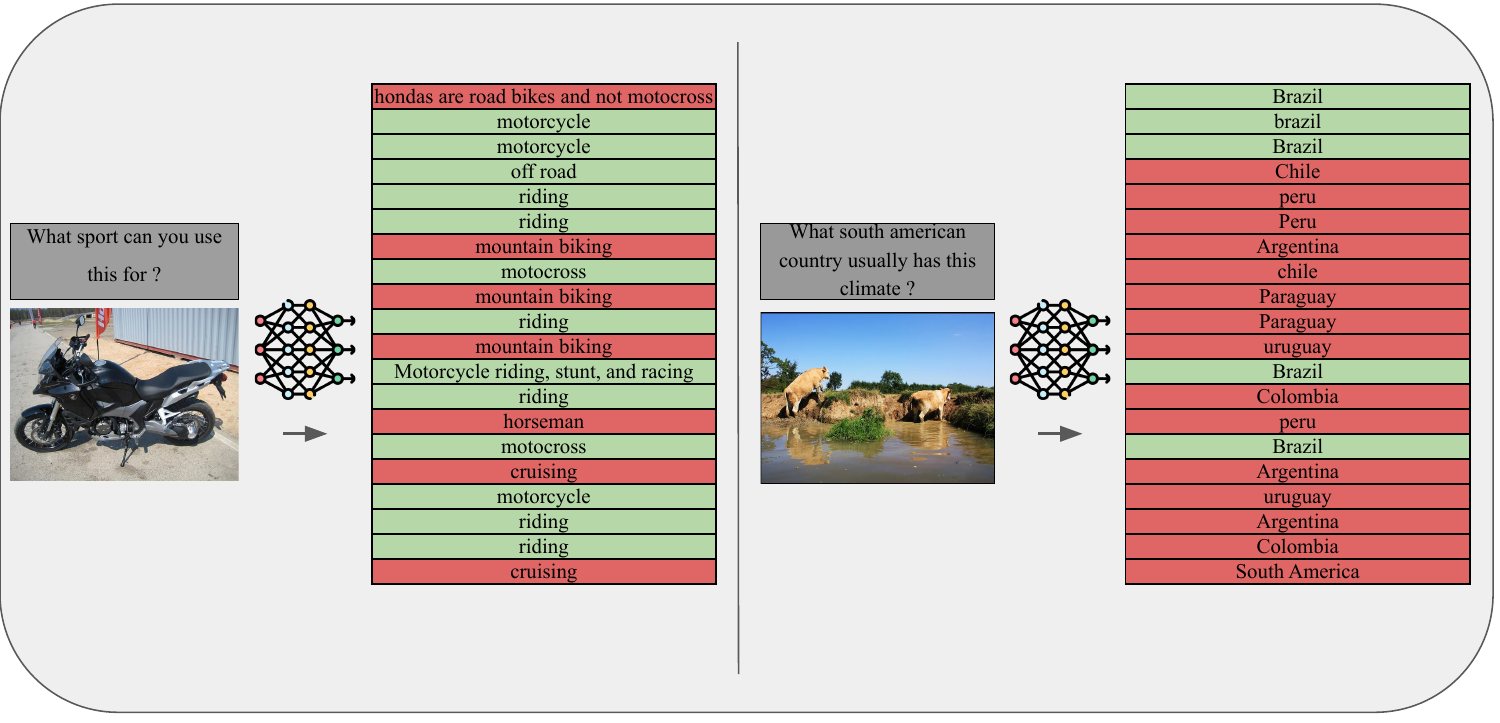}
    \caption{\textbf{Illustration of the truthfulness labeling strategy} using two examples from the \textbf{OKVQA} dataset: one where the truthfulness label $l^{(i)}=1$ and another where $l^{(i)}=0$. Each generated answer is compared to a reference answer using an LLM-as-jugde approach. Answers depicted in \textcolor{green}{green} ($l_j^{(i)}=1$) are correct, whereas those  in \textcolor{red}{red} are incorrect ($l_j^{(i)}=0$).}
    \label{fig:annotations}
\end{figure*}

Let $\vx^{(i)}$ denote a mulimodal input (question+image) and let $\mathcal{Y}^{(i)}:=\{\vy^{(j)}_{i}\}_{j=1}^{20}$ be the set of $20$ candidate answers generated by a LVLM given $\vx^{(i)}$, as described in Section~\ref{sec:SGPU-GP}.
Following a standard protocol in the UQ literature~\cite{band2021, kuhnsemantic, farquhar2024detecting, janiak2025, manakul2023}, we evaluate the correctness $l^{(i)}$ of each generated response $\vy_i^{(j)}$ with respect to a reference answer $\bar \vy_i$ using an LLM-as-judge approach.
This approach ensures a fair comparison with prior work, achieves better performance than other metrics (e.g., ROUGE score), and shows strong agreement with human judgments~\cite{janiak2025}.
We use \texttt{Llama-3.1-8B} on the train set to compare each output $\vy^{(j)}_{i}$ with the reference answer $\bar \vy_{i}$ and assign a binary score $l_j^{(i)} \in \{0,1\}$ to indicate whether $\vy^{(j)}_{i}$ is correct. For the test set, we use a different and bigger model: \texttt{Llama-3.1-70B} in order to break
any potential circular dependency.
In particular, we use the following evaluation prompt:
\begin{tcolorbox}[colback=gray!5,colframe=black!90,title=Prompt for Evaluating Truthfulness of Generated Answers]
    We are assessing the quality of answers to the following question: [\texttt{question}]. \\
    The expected answer is: [\texttt{correct answers}].\\
    The proposed answer is: [\texttt{predicted answer}].\\
    Within the context of the question, does the proposed answer mean the same as the expected answer?\\
    Respond only with yes or no.\\
    Response:
\end{tcolorbox}
\noindent The truthfulness label $l^{(i)}$ that indicate the correctness of generated answers $\mathcal{Y}^{(i)}$ is then defined as the most frequent label among $\{l^{(i)}_{1}, \ldots, l^{(i)}_{20}\}$.
Figure~\ref{fig:annotations} illustrates the labeling strategy with two samples from the \textbf{OKVQA} dataset and candidate answers generated by \texttt{Qwen2.5-VL-3B}. 

Finally, we conducted an experiment varying the Judges used on the test set to verify that \method's performance gains are not dependent on the choice of Judge. Results are summarized in Tab~\ref{tab:judge}, which reports the mean AUROC across all data/model combinations for each method. Across all labeling settings, \method~ consistently outperforms competing methods, further demonstrating its robustness.

\begin{table}[]
\centering
\caption{Mean AUROC varying LLM judges on the eval set.  \textbf{Bold} values indicates best performance, \underline{underline} values indicate the second-best.}
\label{tab:judge}
\resizebox{0.5\textwidth}{!}{
\begin{tabular}{cccc}
\toprule
\multicolumn{1}{l}{} & \texttt{Llama-3.1-70B} & \texttt{Qwen-2.5-32B} & \texttt{Llama-3.3-70B} \\
\multicolumn{1}{l}{} & \textbf{AUROC$\uparrow$} & \textbf{AUROC$\uparrow$} & \textbf{AUROC$\uparrow$} \\
\midrule
SE & 0,755 & 0.762 & 0.760 \\
PE & 0.751 & 0.756 & 0.752 \\
DSE & 0.757 & 0.760 & 0.761 \\
KLE & 0.765 & \underline{0.771} & 0.763 \\
UMPIRE & 0.723 & 0.743 & 0.734 \\
Cov Eigenscore & 0.716 & 0.723 & 0.718 \\
Cos Eigenscore & 0.746 & 0.749 & 0.742 \\
SAPLMA & 0.761 & 0.753 & 0.732 \\
SEP & 0.711 & 0.710 & 0.703 \\
\rowcolor{gray!20}Global \method~(ours) & \underline{0.766} & 0.764 & \underline{0.765} \\
\rowcolor{green!20}\method (ours) & \textbf{0.788} & \textbf{0.784} & \textbf{0.782}\\
\midrule~
\end{tabular}
}
\end{table}

\newpage

\section{Motivations Behind the Gaussian Process}
\label{app_GP_motivation}

As mentioned in Section~\ref{sec:Method}, the use of a Gaussian Process Classifier (GPC) is motivated by its ability to provide direct access to internal uncertainty estimates (through the predictive standard deviation), its efficiency with limited training data, and its good calibration. 

\subsection{\method~Internal Uncertainty}
\label{app_gp_internal_uncertainty}

\begin{figure}[t]
    \centering
    \begin{center}
        \includegraphics[width=0.98\linewidth]{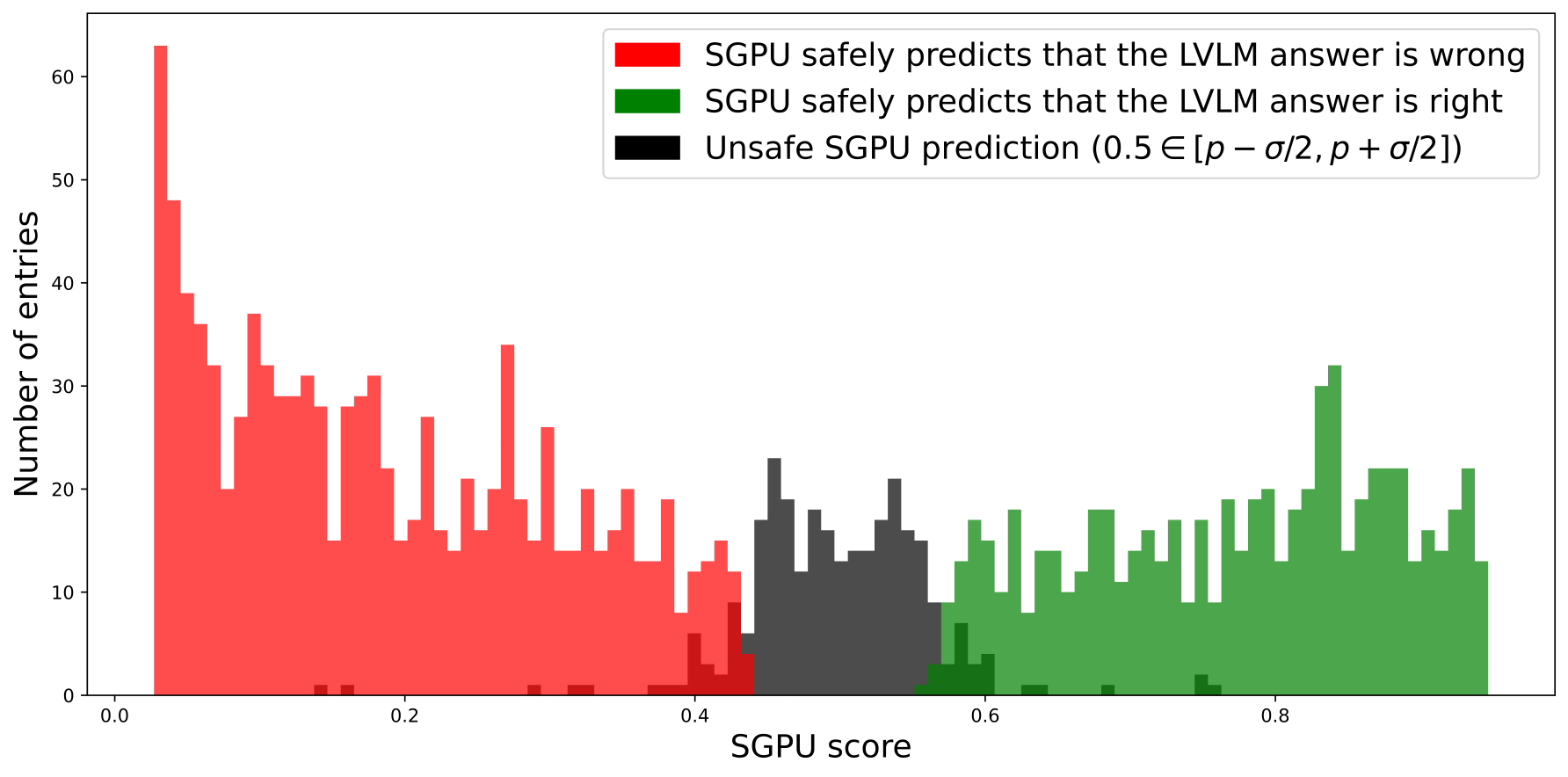}
    \end{center}
    \caption{\textbf{Detection of Unsafe SGPU Predictions} with \texttt{llava-mistral-7b} on \textbf{VIZWIZ}. 
    Unsafe SGPU predictions are defined when $0.5 \in [p-\sigma/2, p+\sigma/2]$, where $p$ is the SGPU score and $\sigma$ is its associated predicted standard deviation.}
    \label{app_fig:gp_histo}
\end{figure}

For each prediction, a GPC provides two values: the class probability and the corresponding predictive standard deviation.
In the SGPU framework, the GPC returns the probability that the generated answers are similar.
Predicted probabilities close to~$1$ or $0$ indicate thus high confidence, whereas values near~$0.5$ correspond to regions of uncertainty. 
A straightforward approach to exclude uncertain predictions is to discard samples whose probabilities are close to $0.5$.
However, \method~ adopts a more principled approach to quantifying predictive uncertainty by leveraging the predictive standard deviation, which captures the internal uncertainty for each prediction. 
A prediction is flagged as \emph{unsafe} if adding or subtracting half of the predictive standard deviation to the predicted probability changes the resulting class label.
Figure~\ref{app_fig:gp_histo} depicts this classification strategy on \textbf{VizWiz} with \texttt{llava-mistral-7b}, where unsafe predictions are highlighted in black. 
To verify that these predictions are indeed unreliable, we compute the AUROC (using the same setup as in the main experiment) before and after removing the unsafe entries. 
As expected, the AUROC increases from $0.848$ to $0.879$ after filtering these entries, confirming that our criterion effectively identifies uncertain predictions.

\newpage
\subsection{\method~Training}
\label{app_gp_training}
Unlike popular semantic UQ methods (Section~\ref{sec:background}), SGPU requires training. 
Nevertheless, Gaussian Processes are well known for their strong performance even with limited training data as illustrated by Table~\ref{tab:training_req}. 
The results show that SGPU maintains competitive performance even with a very small number of training samples.
These findings confirm the applicability and portability of SGPU.

\begin{table}[t]
\centering
\scriptsize
\setlength{\tabcolsep}{5pt}
\caption{\textbf{\method~requires only a few training samples.} 
Evaluation of \method~with varying number of training samples, where \method~$n$ denotes SGPU trained with only $n$ samples. 
\textbf{Bold} values indicate best performance.}
\label{tab:training_req}
\begin{tabular}{c|c|ccc}
\toprule
& & \multicolumn{3}{c}{\texttt{Qwen2.5-VL-3B}} \\
\textbf{Dataset} & \textbf{Method} 
& \textbf{AUROC$\uparrow$} & \textbf{AUARC$\uparrow$} & \textbf{ECE$\downarrow$}\\

\midrule
\multirow{15}{*}{\rotatebox{90}{\textbf{VIZWIZ}}}
& SE & 0.826 & 0.620 & 0.191 \\
& PE & 0.812 & 0.608 & 0.134 \\
& DSE & 0.834 & 0.613 & 0.219 \\
& KLE & 0.834 & 0.622 & 0.255 \\
& UMPIRE & 0.808 & 0.608 & 0.217 \\
& Cos Eigenscore & 0.678 & 0.524 & 0.086 \\
& Cos Eigenscore & 0.807 & 0.606 & 0.156 \\
& SAPLMA & 0.702 & 0.533 & 0.159 \\
 & SEP & 0.697 & 0.533 & 0.169 \\
& \cellcolor{green!20} \textbf{\method} (ours) & \cellcolor{green!20} \textbf{0.858} & \cellcolor{green!20} \textbf{0.635} & \cellcolor{green!20} \textbf{0.037} \\
 & \method~2000      & 0.856 & 0.633 & 0.041 \\ 
 & \method~1000      & 0.853 & 0.632 & 0.044 \\ 
 & \method~300       & 0.851 & 0.632 & 0.037 \\
  & \method~200       & 0.851 & 0.632 & 0.040 \\
 & \method~100       & 0.848 & 0.630 & 0.066 \\ 
\midrule

\end{tabular}
\end{table}

\section{Additional Experiments}
\label{app_additional}
\subsection{Additional Experiments on LLM}\label{app_LLM}
To assess whether Global-\method~ transfers beyond the vision-language setting, we evaluate it zero-shot on two text-only benchmarks, \textbf{TriviaQA} and \textbf{GSM8K}, with \texttt{Llama-3.1-8B-Instruct} and \texttt{Qwen2.5-3B-Instruct} as target models. Zero-shot here means the target model and target dataset are held out entirely: Global-SGPU is trained only on response sets pooled from other source models and datasets, with no exposure to the target architecture or data during training or hyperparameter selection. Table~\ref{tab:results_LLM} reports AUROC, AUARC, and ECE against the SE, DSE, and KLE baselines.

\begin{table}[t]
\centering
\caption{AUROC, AUARC, and ECE across datasets and models for LLM zero shot tranfer.}
\resizebox{0.5\textwidth}{!}{
\begin{tabular}{cccccccc|ccccc}
\toprule
& & \multicolumn{3}{c}{\texttt{Llama-3.1-8B-Instruct}} & \multicolumn{3}{c}{\texttt{Qwen2.5-3B-Instruct}} & \multicolumn{3}{c}{\textbf{Mean}} \\
\midrule
\textbf{Dataset} & \textbf{Method} & AUROC & AUARC & ECE & AUROC & AUARC & ECE & AUROC & AUARC & ECE \\
\midrule
\textbf{TriviaQA} & SE & 0.842 & 0.823 & 0.217 & 0.654 & 0.511 & 0.258 & 0.748 & 0.667 & 0.238 \\
\textbf{TriviaQA} & DSE & 0.853 & 0.827 & 0.243 & 0.660 & 0.504 & 0.265 & 0.757 & 0.666 & 0.254 \\
\textbf{TriviaQA} & KLE & 0.887 & 0.841 & 0.139 & 0.821 & 0.598 & 0.232 & 0.854 & 0.720 & 0.186 \\
\textbf{TriviaQA} & Global-\method~(ours)  & 0.849 & 0.821 & 0.210 & 0.836 & 0.610 & 0.153 & 0.843 & 0.716 & 0.182 \\
\midrule
\textbf{GSM8K} & SE & 0.738 & 0.300 & 0.233 & 0.773 & 0.174 & 0.435 & 0.756 & 0.237 & 0.334 \\
\textbf{GSM8K} & DSE & 0.746 & 0.305 & 0.211 & 0.776 & 0.174 & 0.428 & 0.761 & 0.240 & 0.320 \\
\textbf{GSM8K} & KLE & 0.696 & 0.271 & 0.468 & 0.777 & 0.175 & 0.551 & 0.737 & 0.223 & 0.510 \\
\textbf{GSM8K} & Global-\method~(ours) & 0.720 & 0.295 & 0.286 & 0.778 & 0.175 & 0.424 & 0.749 & 0.235 & 0.355 \\
\bottomrule
\end{tabular}
}
\label{tab:results_LLM}
\end{table}

\newpage
\subsection{Transferability}
\label{app_transfer}

As discussed in Section~\ref{sec:Experiment}, an interesting aspect is that the trained SGPU generalizes well across different LVLM architectures in plug-and-play manner without any adjustement or fine-tuning.
For VQA tasks, Table~\ref{app_tranfer_SGPU} shows substantial cross-model transfer, where a model trained using outputs from one LVLM can be applied effectively to another with only a small loss in performance. 
These results suggest that the distribution of eigenvalues is not strongly dependent on any specific LVLM models, and that our method exhibits robust cross-model generalization.

\begin{table}[]
    \centering
    \caption{\method transfer effectiveness across LVLM architectures on \textbf{OKVQA}.}

    \resizebox{0.5\textwidth}{!}{
    \begin{tabular}{cc|ccc}
    \toprule
    \textbf{Trained on} & \textbf{Tested on} &\textbf{AUROC$\uparrow$} & \textbf{AUARC$\uparrow$} & \textbf{ECE$\downarrow$}\\
    \midrule
    \texttt{llava-mistral-7b} & \texttt{llava-mistral-7b} & 0.821 & 0.810 & 0,109 \\
    \texttt{Qwen2.5-VL-3B} & \texttt{llava-mistral-7b} & 0.816 & 0.807 & 0.083 \\
    \texttt{Qwen2.5-VL-7B} & \texttt{llava-mistral-7b} & 0.805 & 0.804 & 0.072 \\
    \texttt{idefics2-8b-chatty} & \texttt{llava-mistral-7b} & 0.784 & 0.794 & 0.174 \\
    \multicolumn{1}{l}{} & \multicolumn{1}{l}{} & \multicolumn{1}{l}{} & \multicolumn{1}{l}{} & \multicolumn{1}{l}{} \\
    \texttt{Qwen2.5-VL-3B} & \texttt{Qwen2.5-VL-3B} & 0.841 & 0.801 & 0.088 \\
    \texttt{llava-mistral-7b} & \texttt{Qwen2.5-VL-3B} & 0.836 & 0.799 & 0.098 \\
    \texttt{Qwen2.5-VL-7B} & \texttt{Qwen2.5-VL-3B} & 0.814 & 0.790 & 0.092 \\
    \texttt{idefics2-8b-chatty} & \texttt{Qwen2.5-VL-3B} & 0.791 & 0.783 & 0.172 \\
    \multicolumn{1}{l}{} & \multicolumn{1}{l}{} & \multicolumn{1}{l}{} & \multicolumn{1}{l}{} & \multicolumn{1}{l}{} \\
    \texttt{Qwen2.5-VL-7B} & \texttt{Qwen2.5-VL-7B} & 0.725 & 0.786 & 0.182 \\
    \texttt{llava-mistral-7b} & \texttt{Qwen2.5-VL-7B} & 0.722 & 0.784 & 0.274 \\
    \texttt{Qwen2.5-VL-3B} & \texttt{Qwen2.5-VL-7B} & 0.717 & 0.780 & 0.115 \\
    \texttt{idefics2-8b-chatty} & \texttt{Qwen2.5-VL-7B} & 0.704 & 0.774 & 0.164 \\
    \multicolumn{1}{l}{} & \multicolumn{1}{l}{} & \multicolumn{1}{l}{} & \multicolumn{1}{l}{} & \multicolumn{1}{l}{} \\
    \texttt{idefics2-8b-chatty} & \texttt{idefics2-8b-chatty} & 0.712 & 0.755 & 0.153 \\
    \texttt{llava-mistral-7b} & \texttt{idefics2-8b-chatty} & 0.696 & 0.746 & 0.229 \\
    \texttt{Qwen2.5-VL-3B} & \texttt{idefics2-8b-chatty} & 0.708 & 0.753 & 0.111 \\
    \texttt{Qwen2.5-VL-7B} & \texttt{idefics2-8b-chatty} & 0.697 & 0.748 & 0.138 \\
    \bottomrule 
    \end{tabular}
    }
    \label{app_tranfer_SGPU}
\end{table}

\subsection{Beyond self consistency}
\label{app_self_consistency}
Direct self-consistency methods tend to assign high confidence when a model repeatedly produces the same answer, even when that answer is incorrect.
SGPU can identify some of these failures.
On \texttt{Qwen2.5-VL-7B}/\textbf{CIFAR}, among the 4{,}058 response sets in which the same answer appeared at least ten times, SGPU assigned low confidence to 304 response sets that were indeed incorrect.
This shows that SGPU mitigates, rather than completely resolves, the failure of direct self-consistency.
The complete confusion matrix, precision, recall, and qualitative examples can be found in Tab~\ref{tab:self_consistency_full}.

\begin{table}[t]
\centering
\caption{Full confusion matrix for SGPU vs.\ direct self-consistency on response sets with a consistent cluster (a semantic id shared by more than $10$ samples). TN: consistent \& incorrect, correctly flagged uncertain by SGPU. Accuracy: $(\text{TP}+\text{TN})/(\text{TP}+\text{FP}+\text{TN}+\text{FN})$.}
\label{tab:self_consistency_full}
\resizebox{0.5\textwidth}{!}{%
\begin{tabular}{ccccc}
\toprule
Model & Dataset & TN & Recall & Accuracy \\
\midrule
\texttt{llava-mistral-7b} &\textbf{ VIZWIZ} & 0 & 0.953 & 0.902 \\
 & \textbf{VQARAD} & 0 & 0.889 & 0.533 \\
 & \textbf{OKVQA } & 22 & 0.986 & 0.910 \\
 & \textbf{Imagenette} & 680 & 0.831 & 0.649 \\
 &  \textbf{CIFAR } & 224 & 0.905 & 0.885 \\
 & \textbf{ADVQA} & 22 & 0.931 & 0.726 \\
\texttt{Qwen2.5-VL-3B} & \textbf{ VIZWIZ}& 2 & 0.990 & 0.906 \\
 &\textbf{VQARAD}& 0 & 0.956 & 0.670 \\
 & \textbf{OKVQA } & 29 & 0.971 & 0.887 \\
 & \textbf{Imagenette} & 292 & 0.872 & 0.639 \\
 &  \textbf{CIFAR } & 129 & 0.948 & 0.830 \\
 & \textbf{ADVQA} & 34 & 0.858 & 0.746 \\
\texttt{Qwen2.5-VL-7B} &\textbf{ VIZWIZ}& 0 & 0.989 & 0.940 \\
 & \textbf{VQARAD} & 0 & 1.000 & 0.842 \\
 & \textbf{OKVQA }& 13 & 0.946 & 0.896 \\
 & \textbf{Imagenette}& 762 & 0.794 & 0.722 \\
 &  \textbf{CIFAR } & 304 & 0.844 & 0.814 \\
 & \textbf{ADVQA} & 8 & 0.887 & 0.783 \\
\texttt{idefics2-8b-chatty} & \textbf{ VIZWIZ} & 3 & 0.944 & 0.864 \\
 & \textbf{VQARAD} & 18 & 0.738 & 0.611 \\
 & \textbf{OKVQA }& 29 & 0.860 & 0.798 \\
 & \textbf{Imagenette} & 373 & 0.724 & 0.616 \\
 & \textbf{CIFAR }& 210 & 0.660 & 0.654 \\
 & \textbf{ADVQA} & 0 & 0.785 & 0.769 \\
\bottomrule
\end{tabular}%
}
\end{table}

\subsection{Number of samples}
\label{app_n_samples}
In our default \method~ configuration, we use $20$ samples, as this is the most commonly adopted value in the literature, and we apply the same number of samples consistently across all benchmark methods for a fair comparison. To justify this choice, we additionally report in Table~\ref{tab:sgpu_samples} the performance of \method when varying the number of samples on \textbf{OKVQA}, \texttt{Qwen2.5-VL-3B}. We observe a clear improvement in performance as the number of samples increases up to $15$, after which performance stabilizes, with $15$ and $20$ samples yielding comparable results. This confirms that using $20$ samples is a reasonable choice, offering performance on par with the point of stabilization while remaining consistent with common practice.

\begin{table}[]
\centering
\caption{Effect of the number of \method~ samples on \textbf{OKVQA} with \texttt{Qwen2.5-VL-3B}.}
\label{tab:sgpu_samples}
\resizebox{0.4\textwidth}{!}{%
\begin{tabular}{cccc}
\toprule
Number of samples & \textbf{AUROC$\uparrow$} & \textbf{AUARC$\uparrow$} & \textbf{ECE$\downarrow$} \\
\midrule
5 & 0.811 & 0.786 & 0.098 \\
10 & 0.836 & 0.794 & 0.087 \\
15 & 0.842 & 0.800 & 0.089 \\
20 & 0.841 & 0.801 & 0.088 \\
\bottomrule
\end{tabular}
}
\end{table}

\subsection{Statistical Significance}
\label{app_Statistical_Significance}

To assess the statistical significance of our results, we compute 95\%  confidence intervals for the AUROC of \method~ via percentile bootstrap (2000 resamples of the validation predictions), reported alongside the point estimates in Table~\ref{tab:70b_auroc_ci}; the resulting intervals are narrow across most datasets and models (typically $\pm 0.01$–$0.03$), reinforcing the reliability of our reported gains.

\begin{table}[t]
\centering
\caption{AUROCS of \method~ reported with 95\% confidence intervals.}
\resizebox{0.5\textwidth}{!}{
\begin{tabular}{cccc}
\toprule
\textbf{Dataset} & \textbf{Model} & \textbf{AUROC} & \textbf{CI} \\
\midrule
\textbf{VIZWIZ} & \texttt{llava-mistral-7b}& 0.848 & ($\pm$0.011) \\
\textbf{VIZWIZ} & \texttt{Qwen2.5-VL-3B} & 0.858 & ($\pm$0.016) \\
\textbf{VIZWIZ} & \texttt{Qwen2.5-VL-7B} & 0.821 & ($\pm$0.017) \\
\textbf{VIZWIZ} & \texttt{idefics2-8b-chatty} & 0.721 & ($\pm$0.022) \\
\midrule
\textbf{VQARAD} &  \texttt{llava-mistral-7b} & 0.765 & ($\pm$0.048) \\
\textbf{VQARAD} & \texttt{Qwen2.5-VL-3B} & 0.765 & ($\pm$0.047) \\
\textbf{VQARAD} & \texttt{Qwen2.5-VL-7B} & 0.727 & ($\pm$0.046) \\
\textbf{VQARAD} & \texttt{idefics2-8b-chatty} & 0.700 & ($\pm$0.048) \\
\midrule
\textbf{OKVQA} &  \texttt{llava-mistral-7b} & 0.821 & ($\pm$0.011) \\
\textbf{OKVQA} & \texttt{Qwen2.5-VL-3B} & 0.841 & ($\pm$0.011) \\
\textbf{OKVQA} & \texttt{Qwen2.5-VL-7B} & 0.725 & ($\pm$0.014) \\
\textbf{OKVQA} & \texttt{idefics2-8b-chatty} & 0.712 & ($\pm$0.019) \\
\midrule
\textbf{Imagenette} & \texttt{llava-mistral-7b} & 0.788 & ($\pm$0.014) \\
\textbf{Imagenette} & \texttt{Qwen2.5-VL-3B} & 0.775 & ($\pm$0.014) \\
\textbf{Imagenette} & \texttt{Qwen2.5-VL-7B} & 0.813 & ($\pm$0.013) \\
\textbf{Imagenette}& \texttt{idefics2-8b-chatty} & 0.714 & ($\pm$0.016) \\
\midrule
\textbf{CIFAR} &  \texttt{llava-mistral-7b} & 0.944 & ($\pm$0.008) \\
\textbf{CIFAR} & \texttt{Qwen2.5-VL-3B} & 0.848 & ($\pm$0.011) \\
\textbf{CIFAR} & \texttt{Qwen2.5-VL-7B} & 0.919 & ($\pm$0.009) \\
\textbf{CIFAR} & \texttt{idefics2-8b-chatty} & 0.820 & ($\pm$0.011) \\
\midrule
\textbf{ADVQA} &  \texttt{llava-mistral-7b} & 0.747 & ($\pm$0.025) \\
\textbf{ADVQA} & \texttt{Qwen2.5-VL-3B} & 0.749 & ($\pm$0.026) \\
\textbf{ADVQA} & \texttt{Qwen2.5-VL-7B} & 0.686 & ($\pm$0.027) \\
\textbf{ADVQA} & \texttt{idefics2-8b-chatty} & 0.792 & ($\pm$0.292) \\
\bottomrule
\end{tabular}
}
\label{tab:70b_auroc_ci}
\end{table}

\subsection{Complete Results}
\label{app_complete_results}

Table~\ref{tab:summary_exp} presents a summary of all results discussed in Section~\ref{sec:Experiment}.

\begin{table*}[t]
\centering
\scriptsize
\setlength{\tabcolsep}{4pt}
\caption{\textbf{\method~ consistently improves AUROC and AUARC while reducing ECE.} Comparison across vision datasets, metrics, and LVLM backbones.  \textbf{Bold} values indicates best performance, \underline{underline} values indicate the second-best.}
\label{tab:summary_exp}
\resizebox{0.9\textwidth}{!}{
\begin{tabular}{c|c|ccc|ccc|ccc|ccc|ccc}
\toprule
& & \multicolumn{3}{c}{\texttt{llava-mistral-7b}} 
& \multicolumn{3}{c}{\texttt{Qwen2.5-VL-3B}} 
& \multicolumn{3}{c}{\texttt{Qwen2.5-VL-7B}}
& \multicolumn{3}{c}{\texttt{idefics2-8b-chatty}} \\
Dataset & Method 
& AUROC & AUARC & ECE 
& AUROC & AUARC & ECE 
& AUROC & AUARC & ECE 
& AUROC & AUARC & ECE \\
\midrule
\midrule
\textbf{VIZWIZ} & SE & 0.804 & 0.739 & 0.321 & 0.826 & 0.620 & 0.191 & 0.806 & 0.707 & 0.288 & 0.717 & 0.715 & 0.307 \\
\textbf{VIZWIZ} & PE & 0.798 & 0.733 & \textbf{0.103} & 0.812 & 0.608 & 0.134 & 0.798 & 0.698 & \underline{0.064} & 0.692 & 0.699 & \underline{0.095} \\
\textbf{VIZWIZ} & DSE & 0.807 & 0.734 & 0.342 & 0.834 & 0.613 & 0.219 & 0.811 & 0.698 & 0.303 & 0.720 & 0.712 & 0.329 \\
\textbf{VIZWIZ} & KLE & 0.823 & 0.744 & 0.149 & 0.834 & 0.622 & 0.255 & \textbf{0.840} & \textbf{0.723} & 0.180 & \textbf{0.773} & \textbf{0.735} & 0.141 \\
\textbf{VIZWIZ} & UMPIRE & 0.823 & 0.743 & 0.146 & 0.808 & 0.608 & 0.217 & 0.774 & 0.687 & 0.307 & 0.687 & 0.697 & 0.293 \\
\textbf{VIZWIZ} & Cov Eigenscore & 0.754 & 0.689 & 0.201 & 0.678 & 0.524 & \underline{0.086} & 0.703 & 0.646 & 0.159 & 0.637 & 0.660 & 0.274 \\
\textbf{VIZWIZ}& Cos Eigenscore & 0.795 & 0.734 & 0.183 & 0.807 & 0.606 & 0.156 & 0.796 & 0.696 & 0.183 & 0.727 & 0.713 & 0.211 \\
\textbf{VIZWIZ}& SAPLMA & 0.745 & 0.703 & \textbf{0.103} & 0.702 & 0.533 & 0.159 & 0.667 & 0.622 & 0.137 & 0.683 & 0.693 & 0.191 \\
\textbf{VIZWIZ} & SEP & 0.678 & 0.667 & 0.287 & 0.697 & 0.533 & 0.169 & 0.686 & 0.637 & 0.292 & 0.652 & 0.678 & 0.367 \\
\rowcolor{gray!20}\textbf{VIZWIZ} & Global- \method~(ours) & \textbf{0.849} & \underline{0.764} & 0.143 & \underline{0.851} & \underline{0.631} & 0.111 & 0.816 & 0.705 & 0.111 & 0.725 & 0.711 & 0.172 \\
\rowcolor{green!20}\textbf{VIZWIZ} & \textbf{\method } (ours) & \underline{0.848} & \textbf{0.770} & \underline{0.134} & \textbf{0.858} & \textbf{0.635} & \textbf{0.037} & \underline{0.821} & \underline{0.709} & \textbf{0.061} & \underline{0.721} & \underline{0.715} & \textbf{0.127} \\

\midrule

\textbf{VQARAD} & SE & 0.685 & 0.413 & 0.181 & 0.692 & 0.619 & 0.196 & 0.730 & 0.657 & 0.315 & 0.682 & 0.524 & \underline{0.083} \\
\textbf{VQARAD} & PE & 0.719 & 0.427 & \underline{0.108} & 0.770 & 0.662 & 0.078 & 0.732 & 0.647 & \underline{0.076} & 0.601 & 0.480 & 0.211 \\
\textbf{VQARAD} & DSE & 0.683 & 0.407 & 0.190 & 0.696 & 0.619 & 0.214 & 0.731 & 0.651 & 0.328 & 0.678 & 0.522 & 0.086 \\
\textbf{VQARAD} & KLE & 0.578 & 0.368 & 0.346 & 0.721 & 0.655 & 0.150 & 0.751 & 0.677 & 0.208 & 0.694 & 0.523 & 0.339 \\
\textbf{VQARAD} & UMPIRE & 0.687 & 0.410 & 0.206 & 0.685 & 0.606 & 0.196 & 0.711 & 0.640 & 0.348 & 0.627 & 0.489 & 0.121 \\
\textbf{VQARAD} & Cov Eigenscore & 0.685 & 0.456 & 0.234 & 0.744 & 0.636 & 0.254 & 0.710 & 0.645 & 0.348 & 0.676 & 0.527 & 0.088 \\
\textbf{VQARAD} & Cos Eigenscore & 0.675 & \textbf{0.485} & 0.219 & 0.714 & 0.629 & \underline{0.046} & 0.734 & 0.650 & 0.224 & 0.653 & 0.505 & 0.212 \\
\textbf{VQARAD} & SAPLMA & \underline{0.799} & 0.426 & \textbf{0.100} & \underline{0.817} & \underline{0.704} & 0.171 & \textbf{0.803} & \underline{0.692} & 0.121 & \underline{0.720} & \underline{0.593} & 0.141 \\
\textbf{VQARAD} & SEP & \textbf{0.819} & 0.454 & 0.123 & \textbf{0.823} & \textbf{0.753} & 0.181 & \underline{0.793} & \textbf{0.746} & 0.143 & \textbf{0.786} & \textbf{0.746} & 0.147 \\
\rowcolor{gray!20}\textbf{VQARAD} & Global- \method~(ours) & 0.777 & \underline{0.463} & 0.150 & 0.740 & 0.641 & 0.117 & 0.741 & 0.650 & 0.089 & 0.645 & 0.494 & 0.094 \\
\rowcolor{green!20}\textbf{VQARAD} & \textbf{\method } (ours) & 0.765 & 0.458 & 0.149 & 0.765 & 0.661 & \textbf{0.057} & 0.727 & 0.649 & \textbf{0.046} & 0.700 & 0.530 & \textbf{0.062} \\

\midrule

\textbf{OKVQA} & SE & 0.807 & 0.804 & 0.177 & 0.807 & 0.788 & 0.119 & 0.709 & 0.782 & 0.444 & \underline{0.757} & 0.778 & 0.249 \\
\textbf{OKVQA}  & PE & 0.784 & 0.795 & \textbf{0.050} & 0.791 & 0.780 & \textbf{0.057} & 0.674 & 0.764 & 0.221 & 0.687 & 0.745 & 0.082 \\
\textbf{OKVQA}  & DSE & \underline{0.809} & 0.803 & 0.205 & 0.817 & 0.791 & 0.162 & 0.708 & 0.779 & 0.456 & \underline{0.757} & 0.776 & 0.272 \\
\textbf{OKVQA}  & KLE\_HEAT & 0.780 & 0.794 & 0.147 & 0.820 & 0.793 & 0.159 & \underline{0.729} & \underline{0.790} & 0.185 & \textbf{0.779} & \textbf{0.790} & 0.157 \\
\textbf{OKVQA}  & UMPIRE & 0.776 & 0.790 & 0.178 & 0.764 & 0.766 & 0.111 & 0.719 & 0.785 & 0.537 & 0.698 & 0.752 & 0.281 \\
\textbf{OKVQA}  & Cov Eigenscore & 0.711 & 0.757 & 0.097 & 0.644 & 0.704 & 0.467 & 0.702 & 0.775 & 0.240 & 0.638 & 0.716 & 0.187 \\
\textbf{OKVQA}  & Cos Eigenscore & 0.780 & 0.790 & 0.060 & 0.776 & 0.773 & \textbf{0.057} & \underline{0.729} & 0.785 & 0.151 & 0.721 & 0.761 & \textbf{0.071} \\
\textbf{OKVQA}  & SAPLMA & 0.764 & 0.777 & 0.110 & 0.754 & 0.753 & 0.091 & \textbf{0.755} & \textbf{0.795} & \textbf{0.078} & 0.713 & 0.749 & 0.138 \\
\textbf{OKVQA}  & SEP & 0.751 & 0.775 & 0.253 & 0.682 & 0.537 & 0.644 & 0.671 & 0.762 & 0.463 & 0.637 & 0.721 & 0.361 \\
\rowcolor{gray!20}\textbf{OKVQA}  & Global- \method~(ours) & \underline{0.809} & 0.803 & 0.180 & \underline{0.836} & \underline{0.796} & 0.185 & 0.721 & 0.783 & 0.205 & 0.708 & 0.754 & 0.183 \\
\rowcolor{green!20}\textbf{OKVQA}  & \textbf{\method } (ours) & \textbf{0.821} & \textbf{0.810} & \underline{0.109} & \textbf{0.841} & \textbf{0.801} & \underline{0.088} & 0.725 & 0.786 & \underline{0.151} & 0.712 & \underline{0.778} & \underline{0.153} \\

\midrule

\textbf{Imagenette} & SE & 0.685 & 0.423 & 0.361 & 0.732 & 0.589 & 0.272 & 0.593 & 0.578 & 0.248 & 0.630 & 0.526 & \underline{0.122} \\
\textbf{Imagenette} & PE & 0.685 & 0.425 & 0.352 & 0.724 & 0.594 & 0.161 & 0.683 & 0.656 & 0.124 & 0.646 & 0.546 & 0.213 \\
\textbf{Imagenette} & DSE & 0.686 & 0.430 & 0.324 & 0.737 & 0.591 & 0.221 & 0.586 & 0.576 & 0.221 & 0.627 & 0.524 & 0.133 \\
\textbf{Imagenette} & KLE & 0.642 & 0.408 & 0.468 & 0.719 & 0.582 & 0.356 & 0.619 & 0.588 & 0.356 & 0.629 & 0.518 & 0.363 \\
\textbf{Imagenette} & UMPIRE & 0.689 & 0.436 & 0.400 & 0.654 & 0.542 & 0.231 & 0.554 & 0.541 & 0.268 & 0.600 & 0.509 & 0.137 \\
\textbf{Imagenette}& Cov Eigenscore & 0.646 & 0.498 & 0.314 & 0.747 & 0.605 & 0.290 & \textbf{0.835} & \textbf{0.745} & 0.284 & \textbf{0.756} & \textbf{0.626} & 0.465 \\
\textbf{Imagenette} & cos\_sim\_eigenscore & 0.657 & 0.478 & 0.325 & 0.728 & 0.611 & 0.272 & 0.687 & 0.653 & 0.243 & \underline{0.725} & 0.604 & 0.228 \\
\textbf{Imagenette} & SAPLMA & \underline{0.750} & 0.505 & 0.275 & \textbf{0.806} & \textbf{0.654} & 0.350 & \underline{0.826} & \underline{0.734} & 0.373 & 0.718 & 0.606 & 0.423 \\
\textbf{Imagenette} & SEP & 0.609 & \textbf{0.619} & 0.342 & 0.579 & 0.536 & 0.208 & 0.455 & 0.449 & 0.323 & 0.562 & 0.538 & 0.363 \\
\rowcolor{gray!20}\textbf{Imagenette} & Global- \method~(ours) & 0.709 & 0.464 & \underline{0.146} & 0.748 & 0.610 & \textbf{0.066} & 0.702 & 0.661 & \underline{0.059} & 0.672 & 0.557 & 0.125 \\
\rowcolor{green!20}\textbf{Imagenette} & \textbf{\method } (ours) & \textbf{0.788} & \underline{0.520} & \textbf{0.098} & \underline{0.776} & \underline{0.636} & \textbf{0.066} & 0.813 & 0.727 & \textbf{0.077} & 0.715 & \underline{0.606} & \textbf{0.055} \\

\midrule

\textbf{CIFAR} & SE & \underline{0.945} & 0.909 & 0.055 & 0.790 & 0.728 & \underline{0.035} & \textbf{0.931} & 0.876 & 0.112 & 0.856 & 0.882 & 0.152 \\
\textbf{CIFAR} & PE & 0.942 & 0.909 & 0.076 & 0.801 & 0.727 & 0.120 & 0.915 & 0.869 & 0.134 & 0.863 & 0.883 & \underline{0.135} \\
\textbf{CIFAR} & DSE & 0.943 & 0.909 & 0.054 & 0.802 & 0.734 & 0.066 & \underline{0.928} & 0.875 & 0.079 & 0.859 & 0.883 & 0.178 \\
\textbf{CIFAR} & KLE & 0.925 & 0.905 & 0.083 & \underline{0.862} & \underline{0.765} & \textbf{0.035} & 0.925 & 0.875 & 0.094 & \textbf{0.907} & \textbf{0.895} & \textbf{0.066} \\
\textbf{CIFAR}& UMPIRE & 0.934 & 0.907 & 0.055 & 0.713 & 0.688 & \underline{0.036} & 0.814 & 0.832 & 0.146 & 0.854 & 0.884 & 0.188 \\
\textbf{CIFAR} & Cov Eigenscore & 0.923 & 0.897 & 0.076 & 0.687 & 0.668 & 0.328 & 0.893 & 0.858 & 0.493 & 0.825 & 0.873 & 0.725 \\
\textbf{CIFAR} & Cos Eigenscore & 0.924 & 0.901 & 0.089 & 0.801 & 0.730 & 0.062 & 0.917 & 0.871 & 0.094 & \underline{0.895} & \underline{0.892} & 0.159 \\
\textbf{CIFAR} & SAPLMA & 0.938 & 0.907 & \textbf{0.045} & \textbf{0.898} & 0.727 & 0.131 & 0.849 & 0.839 & 0.099 & 0.689 & 0.837 & 0.706 \\
\textbf{CIFAR}& SEP & \textbf{0.958} & \textbf{0.932} & \underline{0.046} & 0.803 & \textbf{0.785} & 0.126 & 0.908 & \textbf{0.899} & \textbf{0.050} & 0.840 & 0.709 & 0.282 \\
\rowcolor{gray!20}\textbf{CIFAR} & Global- \method (ours) & 0.921 & 0.899 & 0.279 & 0.839 & 0.752 & 0.175 & 0.894 & 0.858 & 0.240 & 0.852 & 0.876 & 0.233 \\
\rowcolor{green!20}\textbf{CIFAR} & \textbf{\method } (ours)  & 0.944 & 0.909 & 0.118 & 0.848 & 0.759 & 0.072 & 0.920 & 0.874 & 0.136 & 0.820 & 0.868 & 0.364 \\

\midrule

\textbf{ADVQA} & SE & 0.758 & 0.653 & 0.053 & 0.766 & 0.744 & \textbf{0.045} & 0.697 & 0.757 & 0.218 & 0.708 & 0.563 & 0.246 \\
\textbf{ADVQA} & PE & \textbf{0.762} & \textbf{0.655} & 0.221 & 0.762 & 0.743 & 0.072 & 0.668 & 0.741 & \textbf{0.035} & 0.708 & 0.563 & 0.297 \\
\textbf{ADVQA} & DSE & \textbf{0.762} & \underline{0.654} & \textbf{0.035} & \textbf{0.767} & 0.744 & \underline{0.056} & \underline{0.701} & \underline{0.760} & 0.241 & 0.708 & 0.563 & 0.322 \\
\textbf{ADVQA} & KLE& 0.752 & 0.650 & 0.155 & \textbf{0.767} & \textbf{0.745} & 0.111 & \textbf{0.733} & \textbf{0.778} & \underline{0.091} & 0.750 & 0.507 & 0.345 \\
\textbf{ADVQA} & UMPIRE & 0.727 & 0.637 & 0.161 & 0.659 & 0.684 & 0.164 & 0.652 & 0.732 & 0.251 & 0.750 & 0.538 & 0.319 \\
\textbf{ADVQA} & Cov Eigenscore & 0.656 & 0.590 & 0.268 & 0.577 & 0.637 & 0.342 & 0.604 & 0.708 & 0.507 & 0.745 & 0.556 & \textbf{0.186} \\
\textbf{ADVQA} & Cos Eigenscore & 0.717 & 0.632 & 0.204 & 0.599 & 0.654 & 0.078 & 0.596 & 0.700 & 0.149 & 0.756 & \underline{0.593} & \underline{0.190} \\
\textbf{ADVQA}& SAPLMA & 0.728 & 0.635 & 0.144 & 0.600 & 0.658 & 0.150 & 0.591 & 0.700 & 0.292 & \textbf{0.958} & \textbf{0.672} & 0.277 \\
\textbf{ADVQA} & SEP & 0.733 & 0.640 & 0.261 & 0.606 & 0.656 & 0.093 & 0.581 & 0.700 & 0.241 & 0.750 & 0.574 & 0.283 \\
\rowcolor{gray!20}\textbf{ADVQA} & Global- \method~ (ours) & 0.730 & 0.632 & 0.057 & 0.703 & 0.706 & 0.106 & 0.658 & 0.727 & 0.110 & 0.750 & 0.574 & 0.382 \\
\rowcolor{green!20}\textbf{ADVQA}& \textbf{\method } (ours) & 0.747 & 0.645 & \underline{0.039} & 0.749 & 0.734 & 0.108 & 0.687 & 0.746 & 0.188 & \underline{0.792} & 0.558 & 0.225\\

\midrule

\end{tabular}}
\end{table*}

\clearpage

\end{document}